\title{VISUAL-HINT BOUNDARY TO SEGMENT ALGORITHM FOR IMAGE SEGMENTATION}
\author{
        Yu Su and Margaret H. Dunham \\
                Department of Computer Science and Engineering\\
Southern Methodist University\\
Dallas, Texas 75275--0122\\
}
\date{\today}

\documentclass[12pt]{article}
\usepackage{StyleFiles/watermark}

\usepackage[ruled,vlined]{StyleFiles/algorithm2e}
\usepackage{StyleFiles/algorithmic}
\usepackage{amsmath, amsthm, amssymb}
\usepackage{multirow}
\usepackage{thumbpdf}
\usepackage{graphicx}     
\usepackage{wrapfig}     
\usepackage{rotating}    
\usepackage{booktabs}
\usepackage{caption}
\usepackage{subfig}
\usepackage{boxedminipage}  
\usepackage{StyleFiles/qtree}      
\usepackage{fancyhdr}
\usepackage{epsf}
\usepackage{flafter}     

\newtheorem{mydef}{Definition}
\newtheorem{theorem}{Theorem}

\begin{document}
\maketitle

\begin{abstract}
Image segmentation has been a very active research topic in image analysis area. Currently, most of the image segmentation algorithms are designed based on the idea that images are partitioned into a set of regions preserving homogeneous intra-regions and inhomogeneous inter-regions. However, human visual intuition does not always follow this pattern. A new image segmentation method named Visual-Hint Boundary to Segment (VHBS) is introduced, which is more consistent with human perceptions. VHBS abides by two visual hint rules based on human perceptions: (i) the global scale boundaries tend to be the real boundaries of the objects; (ii) two adjacent regions with quite different colors or textures tend to result in the real boundaries between them. It has been demonstrated by experiments that, compared with traditional image segmentation method, VHBS has better performance and also preserves higher computational efficiency.

\end{abstract}

\section{Introduction}

Image segmentation is a vast topic in image analysis. In this chapter, we present a low-level image segmentation method, which has been proposed to segment images in a way that agrees with human perceptions. In recent years, Most of the image segmentation algorithms are designed based on an idea that partitions the images into a set of regions preserving homogeneous intra-regions and inhomogeneous inter-regions. By this idea, these methods segment images in classification or clustering manner. However, human visual intuition does not always follow this manner. Our goal of this research is to define a low-level image segmentation algorithm which is consistent with human visual perceptions.

The proposed new image segmentation method is called Visual-hint Boundary to Segment (VHBS). VHBS abides by two visual hint rules based on human perceptions: (i) the global scale boundaries tend to be the real boundaries of the objects; (ii) two adjacent regions with quite different colors or textures tend to result the real boundaries between them. Compared with other unsupervised segmentation methods, the outputs of VHBS are more consistent to the human perceptions. Beside, reducing complexity is another objective of VHBS since high performance segmentation methods usually are computationally intensive. Therefore, chaos and non-chaos concepts are introduced in VHBS to prevent algorithm going down to details of pixel level. It guarantees that segmentation process stays at a coarse level and keeps the computational efficiency.

VHBS is composed by two phases, Entropy-driven Hybrid Segmentation (EDHS) and Hierarchical Probability Segmentation (HPS.) VHBS starts from EDHS, which produces a set of initial segments by combining local regions and boundaries. These local regions and boundaries are generated by a top-down decomposition process and initial segments are formed by a bottom-up composition process. The top-down decomposition process recursively decomposes the given images into small partitions by setting a stopping condition for each branch of decomposition. We set an entropy measurement as the stopping condition since smaller entropy of local partitions implies lower disorder in the local partitions. To preserve the computational efficiency, we set up a size threshold of the partitions to prevent the decomposition going down to pixel level. Based on this threshold, local partitions are grouped into two types, chaos if the size of a partition is less than the threshold and non-chaos otherwise. Local regions and boundaries are computed in local partitions. Each local region is described by a vector called feature description and the local boundaries are weighted by the probabilities. To calculate the probabilities, we design two scale filter, $f_1$ and $f_2$, which are based on the two visual hints (i) and (ii) respectively. The boundaries between two adjacent regions are weighted by the product of $f_1$ and $f_2$. A bottom-up composition process is followed and combines these local regions and boundaries to form a set of the initial segments, $S=(s_0,\ldots,s_n)$.

The second phase of VHBS is Hierarchical Probability Segmentation (HPS,) which constructs a Probability Binary Tree (PBT) based on these initial segments $S$. PBT presents the hierarchy segments based on boundary probabilities between these initial segments, which forms the leaves of PBT. The root represents the original images and the intern nodes of PBT are the segments combined by their children. Links are labeled by the boundary probabilities. PBT can be visualized in number of segments or even provides the local details. The difference compared with the methods based on MST such as \cite{Felzenszwalb1998, Haxhimusa2004, Felzenszwalb2004} is that these methods generate the tree structure based on the similarities between pixels. Whereas, our method generate the tree structure based on probabilities between regions. It makes the algorithm insensitive to the noise and it greatly reduces the computational complexity. A similar approach is proposed by \cite{Arbelaez2009}. Compared with this approach, VHBS is more efficient since VHBS prevents the decomposition process going down to pixel level by setting a chaos threshold. The novel aspects of VHBS include:

\begin{enumerate}
 \item Visual-Hint: Algorithm abides by two visual hint rules which force the outputs of VHBS are more consistent to human perceptions;
 \item feature Detection: VHBS outputs a set of feature descriptors, which describe the features for each segment;
 \item computational Efficiency: VHBS has high computational efficiency since the algorithm does not go down to pixel level;
 \item hybrid algorithm: VHBS combines edge-, region-, cluster- and graph-based techniques.
\end{enumerate}

\section{Relative work}

Image segmentation is one of major research areas in image analysis and is already explored for many years. Regularly, segmentation methods partition the given images into a set of segments, where a segment contains a set of pixels that preserve high similarity within a segment and maximize differences between different segments. Some examples of classical image segmentation algorithms are k-means clustering \cite{MacQueen1967}, histogram threshold, region growth and watershed transformation. These methods are efficient and easy to understand but with obvious weaknesses which become barriers for applications. These weaknesses include sensitivity of image noise and textures; improperly merging two disjoint areas with the similar gray values by histogram threshold methods; improper initial condition resulting in incorrect outputs and tending to produce excessive over-segmentation by watershed transformation methods \cite{Hill2003}. All these examples demonstrate that image segmentation is an extensive and difficult research area. In recent years, numerous image segmentation methods have been proposed and greatly overcome those weaknesses. Commonly, segmentation algorithms fall into one or more than one of the following categories: edge-based, region-based, cluster-based and graph-based.

The idea of edge-based segmentation methods is straightforward. Contours and segments highly correlate each other. The closed contours give the segments and segments automatically produce the boundaries. Edge-based segmentation methods rely on contours located in images and then these contours produce the boundaries of the segments. Therefore, much research has focused on contour detection. The classical approaches to detect the edges are to look for the discontinuities of brightness such as in Canny Edge Detection \cite{Canny1986}. \cite{Martin2004} demonstrates that these approaches by looking for discontinuities of brightness are inadequate models for locating the boundaries in natural images. One reason is that texture is a common phenomenon in natural images and these textures produce some unwanted edges. Another reason is that to locate segments in an image by edges requests closed boundaries. These approaches usually provide incontinuous contours, which is inadequate for locating the segments. In recent years, many high performance contour detections have been proposed such as \cite{Ren2008,Martin2004,Mairal2008,Zhu2007a,Maire2008}. One category of contour detections is locating the boundary of an image by measuring the local features. To improve the edge detection performance on the natural images, some approaches consider one or combine more descriptors for each pixel in several feature channels over a different scales and orientations to locate boundaries. \cite{Martin2004} proposes a learning schema approach that considers brightness, color and texture features at each local position and uses a classifier to combine these local features. Based on the research of \cite{Martin2004,Maire2008} combines the spectral component to form the so called globalized probability of boundary to improve the accuracy of the boundary detection. There are many boundary detection and segmentation methods which are oriented energy approach \cite{Morrone1988}. Examples of these approaches are \cite{Ma2000,Perona1991}. To achieve high accuracy, these approaches usually combine more than one local feature channels. The computational complexity becomes a bottleneck if the application requests high computational efficiency. Of course there are many other proposed boundary detection and segmentation algorithms based on rich texture analysis such as \cite{Bovik1990,Jain1991,Ferrari2008}. However, highly accurate image contour and segment detection methods are computationally intensive \cite{Arbelaez2009}. 

To locate the segments from the contours, the contours must form closed regions. An example of such research is \cite{Sumengen2005} by bridging the disconnecting contours or contours tracking to locate the regions. Another recent research is \cite{Arbelaez2009}, which can be divided into two phases: (i) Oriented Watershed Transform (OWT) produces a set of initial regions from a contour detection. Paper selects \textit{gPb} proposed by \cite{Maire2008} as the contour detection algorithm since this contour detector gives high accuracy by the benchmark of BSDB \cite{Martin2001}; (ii) Ultrametric Contour Map (UCM) constructs the hierarchical segments. A tree is generated by a greedy graph-based region merging algorithm, where the leaves are those initial regions and the root is the entire images. The segmentation algorithm proposed by \cite{Arbelaez2009} has high accuracy segmentation performance. But the disadvantage is obvious. \textit{gPb} is a expensive contour detection and \textit{gPb} provides fine gradient initial regions. It can be proved that the time complexity of constructing hierarchical segments over such a fine gradient is also computationally intensive. Other examples of recent contour-segment researches are \cite{Ferrari2008,Ferrari2006}.

Typically, a region-based algorithm is combined with clustering techniques to assemble the sub-regions into final segments and numerous methods fall into these two schemas such as \cite{Tobias2002,Kuan2008,Cheng2002,Chuang2006,Chuang1998,Grady2006,Comaniciu2002}. The common used techniques include region growth, split-merge, supervised and unsupervised classification. \cite{Grady2006} proposes a region growth algorithm, called random walk segmentation, which is a multi-label, user interactive image segmentation. This algorithm starts from a small number of seeds with user-predefined labels. Random walk algorithm can determine the probabilities by assuming a random walker starting at each unlabeled pixel that will first reach one of these user-predefined seeds. By this assumption, pixels are assigned to the label which is the greatest probability based on a random walker. Mean shift \cite{Comaniciu2002} is a well known density estimation cluster algorithm and been widely used for image segmentation and object tracking. Based on the domain probability distribution, the algorithm iteratively climbs the gradient to locate the nearest peak. \cite{Comaniciu2002} demonstrates that mean shift provides good segmentation results and is suitable for real data analysis. But the quadratic computational complexity of the algorithm is a disadvantage and the choice of moving window size is not trivial.

In recent years, much research has been built based on graph theoretic techniques. It has been demonstrated by \cite{Shi1997,Felzenszwalb1998,Ding2008,Haxhimusa2004,Falcao2004,1026237,Ding2006,Felzenszwalb2004,Zabih2004,Boykov2001,Ng2001,Weiss1999,Chang2005,Makrogiannis2005,Estrada2004} that these approaches support image segmentation as well. As pointed out by \cite{Haxhimusa2004}, graph-based segmentation could be roughly divided into two groups. One is tree-structure segmentation and another is graph-cut segmentation. Assuming a 2D image as space $P$, both of these two approaches view $P$ as the collection of a set of subgraphs $<p_1,\ldots,p_n>$, where each $p_i$ is an undivided partition, $P=\bigcup p_i$ and $\phi=p_i \bigcap p_j$ for all $1 \leq i\neq j\leq n$. Commonly, $p_i$ denotes a pixel of the images. Tree-structure \cite{Felzenszwalb1998,Ding2008,Haxhimusa2004,Falcao2004,1026237,Ding2006,Felzenszwalb2004} expresses the split-merge process in a hierarchical manner. The links between parents and children indicate the including relationship and the nodes of the tree denote the pieces of subgraphs. Graph-cut \cite{Zabih2004,Boykov2001,Ng2001,Weiss1999,Chang2005,Makrogiannis2005,Estrada2004} views each element of $<p_1,\ldots,p_n>$ as a vertex and the edges of the graph are defined by the similarities between these adjacent vertices. This process forms a weighted undirected graph $G=(V, E)$ and relies the graph cutting to process the graph partition. 

A common tree-structure approach is minimum spanning tree (MST) \cite{Pop2002}. \cite{Felzenszwalb1998,Felzenszwalb2004} propose an algorithm based on MST. That is using the local variation of intensities to locate the proper granularity of the segments based on the so called Kruskal's minimum spanning tree (KMST). Another recent example of tree-structure approach is \cite{Ding2008}. The purpose of this approach is to find the semantic coherence regions based on $\epsilon$-neighbor coherence segmentation criterion by a so called connected coherence tree algorithm (CCTA). Rather than generating tree based on the pixel similarities, \cite{Arbelaez2009} generate a tree structure based on the region similarities. Tree structure based on region similarities should provide better computational complexity than the structure based on the pixel similarities since $|V|$ is greatly reduced by replacing pixels by regions. 

Graph-cut approaches are also called spectral clustering. The basic idea of the graph-cut approach is partitioning $G=(V, E)$ into disjoint subsets by removing the edges linking subsets. Among these approaches, the normalized cut (Ncut) \cite{Shi1997} is widely used. Ncut proposed a minimization cut criterion which measures the cut cost as a fraction of the total edge connection to all the nodes in the graph. This paper demonstrates that minimizing the normalized cut criterion is equivalent to solving a generalized eigenvector system. Other recent examples of graph-cut approaches are \cite{Zabih2004,Boykov2001}. Graph-cut approaches have been proved to be NP-complete problems and the computational complexity is expensive. These disadvantages become the main barriers for graph-cut methods.

\section{Entropy-driven Hybrid Segmentation}

Entropy-driven Hybrid Segmentation, EDHS, begins with a top-down decomposition from the original input images and finishes with a bottom-up composition. Top-down decomposition quarterly partitions a given image and correspondly produces a quadtree based on a stopping condition. EDHS uses an edge detector, such as Canny Detector \cite{Canny1986}, to locate the boundaries between the local regions in the leaves. These boundaries are weighted by the probabilities computed based on the two visual hint rules.

Bottom-up composition recursively combines the local regions when the two adjacent local regions share a boundary with zero probability. This process forms the initial segments, $S=(s_0,\ldots,s_n)$, and a set of probabilities, $C_b =\{c_{i,j}\}$, which describes the weights of the boundaries between each pair of the adjacent initial segments, where index $i$ and $j$ imply two initial segments $s_i$ and $s_j$, which share a boundary valued by a real number $c_{i,j}\in[0, 1]$, $0\leq i\neq j\leq n$. For each initial segment $s_i$, a feature vector, $fd_i=<v_1,\ldots,v_m>$, is generated to describe this segment. The feature descriptor, such as the $CF$ in BIRCH \cite{TZhange96}, summarizes the important features of each area (cluster.) Although the specific values used in feature descriptor may vary, in this chapter we assume $<r, g, b>$, where $r$, $g$ and $b$ are mean values of color channels of red, green and blue.

\subsection{Top-down decomposition}

\begin{figure}[!htb]
  \begin{center}
   \includegraphics[width=4in]{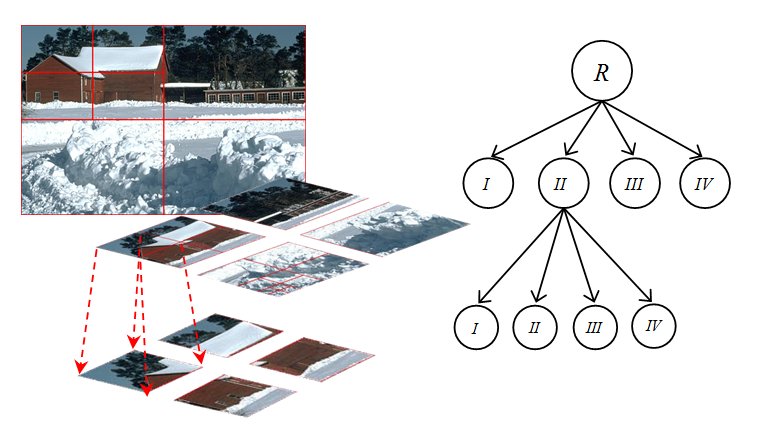}
  \end{center}
  \caption{Top-down decomposition and the quadtree structure} \label{fig:tddecomposition}
\end{figure}

Decomposition mechanism is a wildly used technique in hierarchical image segmentation \cite{Beaulieu1989,Najman1996}. Our decomposition process recursively decomposes the images into four quadrants. The decomposition process is presented by an unbalance quadtree. The root represents the original image and nodes represent the four partitions. A stopping condition is assigned for each branch of decomposition. Partition process is stopped when the desired stopping condition is reached. Figure \ref{fig:tddecomposition} demonstrates an example of the data structures. We summarize the top-down decomposition as follows: 

\renewcommand{\theenumi}{\roman{enumi}}
\renewcommand{\labelenumi}{\theenumi}

\begin{enumerate}
 \item Partitioning the images into small pieces reduces the information presented in local images, which helps VHBS conquer the sub-problems at the local position;
 \item Decomposition provides the relative scale descriptor for the scale filter to calculate the probabilities of the boundaries. We will discuss the relative scale descriptor in section \ref{sec:scalefilter};
 \item Divide and conquer schema potentially supports the parallel implementation, which could greatly improve the computational efficiency.
\end{enumerate}

To describe the decomposition, the dyadic rectangle \cite{Janson2002} is introduced. A dyadic rectangle is the family of regions $I_{s,t}=\{[i2^s,(i+1)2^s-1] \times [j2^t,(j+1)2^t-1], 0\leq i\leq n/2^s-1, 0\leq j\leq m/2^t-1\}$, for $0\leq s\leq \log n, 0\leq t\leq \log m$. The dyadic rectangle of $I$ has some nice properties. Every dyadic rectangle is contained in exactly one ``parent'' dyadic rectangle, where the ``child'' dyadic rectangles are located inside of a ``parent'' dyadic rectangle. The area of ``paren'' is always an integer power of two times of ``chil'' dyadic rectangle. Mapping the images into Cartesian plane, dyadic rectangle provides a model to uniformly decompose images into sub-images recursively.

Given an $n\times m$ $I$, assuming that $n$ and $m$ are the power of $2$, the set of dyadic rectangles at levels $(0, 0)$ through $(\log n, \log m)$ form a complete quadtree, whose root is the level $(\log n, \log m)$ dyadic rectangle $[0, n-1] \times [0, m-1]$. Each dyadic rectangle $I_{s,t}$ with level $1 \leq s \leq \log n, 1 \leq t \leq \log m$ has four children that are dyadic rectangles at levels $s-1$ and $t-1$, which are four quadrants of the $I_{s,t}$. Suppose $I_{s,t}=[i2^s,(i+1)2^s-1] \times [j2^t,(j+1)2^t-1]$, for $0 \leq i \leq n2^{-s}, 0 \leq j \leq m2^{-t}$, then, the first quadrant of $I_{s,t}$ is $[(2i+1)n2^{-(s+1)}, (2i+2)n2^{-(s+1)}-1] \times [2jm2^{-(t+1)},(2j+1)m2^{-(t+1)}-1]$; the second quadrant is $[2in2^{-(s+1)},(2i+1)n2^{-(s+1)}-1] \times [2jm2^{-(t+1)},(2j+1)m2^{-(t+1)}-1]$; the third quadrant is $[2in2^{-(s+1)},(2i+1)n2^{-(s+1)}-1] \times [(2j+1)m2^{-(t+1)},(2j+2)m2^{-(t+1)}-1]$ and the fourth quadrant is $[(2i+1)n2^{-(s+1)},(2i+2)n2^{-(s+1)}-1] \times [(2j+1)m2^{-(t+1)},(2j+2)m2^{-(t+1)}-1]$. 

\subsubsection{Stopping Condition}

In information theory, entropy is a measure of the uncertainty associated with a random variable \cite{Cover1991}. We choose entropy \cite{Shannon2001} as the stopping condition for the top-down decomposition since entropy provides a measurement of disorder of a data set. Let $\zeta$ denote the stopping condition for each branch of the quadtree. If $\zeta$ holds, then EDHS stops the partition process of this branch. By decreasing the size of images, the decomposition reduces the information presented in the local positions. Follows give the concept of segment set. Based on this concept, we define the entropy of images and K-Color Rule. 

\begin{mydef}[Segment set:]\label{def:segments} 
Given a partition $P$ of the interval $[a, b], a=j_0\leq j_1 \leq\ldots\leq j_n=b$, where $a$ and $b$ are minimum and maximum of feature values of a given image $I$, it gives a segment set $X=\{x_1,\ldots,x_k\}$, where $x_i$ is a set of pixels that all the pixels in $x_i$ form a connected region and all the feature values of the pixels are located in interval $[j_h,j_{h+1}], 0\leq h\leq n$. $I=\bigcup x_i$ and $\phi=x_i\bigcap x_j$ for all $0\leq i\neq j\leq k$. 
\end{mydef}

\begin{mydef}[Entropy :]\label{def:entropy} 
Given a segment set $X=\{x_1,\ldots,x_k\}$ based on a partition $P$, then entropy of $I$ to the base $b$ is 
 
 \begin{equation}
 H_k(X)=-\sum_{i=1}^{k}p(x_i)\log_b p(x_i) \label{fun:entropy1}
 \end{equation}
 
\noindent where $p$ denote the probability mass function of segment set $X=\{x_i\}, 1\leq i\leq k$. To make the analysis simple, assume the logarithm base is $e$. This gives 

 \begin{equation}
 H_k(X)=-\sum_{i=1}^{k}p(x_i)\ln p(x_i) \label{fun:entropye}
 \end{equation}

\end{mydef}

\begin{theorem} \label{thr:entropymax} 
  Let $k$ be the number of segments of an image, supremum of $H_k(X)$ is strictly increasing function with respect to $k$ and the supermum of $H_k(X)$ is $\ln(k)$.
\end{theorem}

\noindent\begin{proof}

		Let $H_k(p_1,\ldots,p_k)$ denote the entropy of an image with segment set $X=\{x_1,x_2,\ldots,x_k\}$ and $P(X=x_i)=p_i$. To show supremum of $H_k(X)$ is strictly increasing function respect to $k$, we need to show that $\sup H_k(p_1,\ldots,p_k)\leq \sup H_{k+1}(p'_1,\ldots,p'_{k=1})$ for any $k\in Z^+$.

By \cite{Cover1991} Theorem 2.6.4, we have
\begin{align*} 
		&H_k(p_1,\ldots,p_k)\leq H_k(\frac{1}{k},\ldots,\frac{1}{k}) \text{ and } H_{k+1}(p'_1,\ldots,p'_{k+1})\leq H_{k+1}(\frac{1}{k+1},\ldots,\frac{1}{k+1})\\
    &H_k(\frac{1}{k},\ldots,\frac{1}{k})=-\sum_{i=1}^{k}\frac{1}{k}\ln{1}{k}=\ln(k)\\
		&H_{k+1}(\frac{1}{k+1},\ldots,\frac{1}{k+1})=-\sum_{i=1}^{k+1}\frac{1}{k+1}\ln{1}{k+1}=\ln(k+1)\\
		&\text{Then we have}\\		
		&H_{k+1}(\frac{1}{k+1},\ldots,\frac{1}{k+1})=\ln(k+1)>\ln(k)=H_k(\frac{1}{k},\ldots,\frac{1}{k})
\end{align*} 
\end{proof}

\begin{mydef}[K - Color Rule:]\label{def:kcr} 
Using different colors for different segments in segment set $X=\{x_1,\ldots,x_k \}$,  if the image holds no more than $k$ segments, which means image can be covered by $k$ colors, we say condition `K - Color Rule' (K-CR) is true; else, K-CR is false.
\end{mydef}

Assume an $m \times n$ image $I$. If $I$ is a one color (1-CR) image, then it is a zero entropy image by Theorem \ref{thr:entropymax}. Consider another case. Assume the image is too complicated that none of the segments holds more than one pixel. This case gives the maximum entropy, $\ln(mn)$ ($k$ yields $mn$.) Then, the range of $H_k(X)$ for a $m \times n$ $I$ is $[0, \ln(mn)]$. The larger the entropy is, the more information is contained in the images.

Based on this observation of $H_k(X)$, we choose image entropy as the stopping condition for the top-down decomposition because $H_k(X)$ is highly related to the number of segments. $k$ denotes the number of segments and the range of $k$ is $[1, mn]$. If a proper value of $k$ is chosen for a given image, then $\zeta$  yields to $H_k(X)\leq \ln(k)$ by Theorem \ref{thr:entropymax}. That is, for a certain branch of the quadtree, decomposition approach partitions the given images until the local entropy is no larger than $\ln(k)$.

\subsubsection{K as An Algorithm Tuning Parameter}\label{sec:valueofk}

\begin{figure}[!htb]
  \begin{center}
   \includegraphics[width=4in]{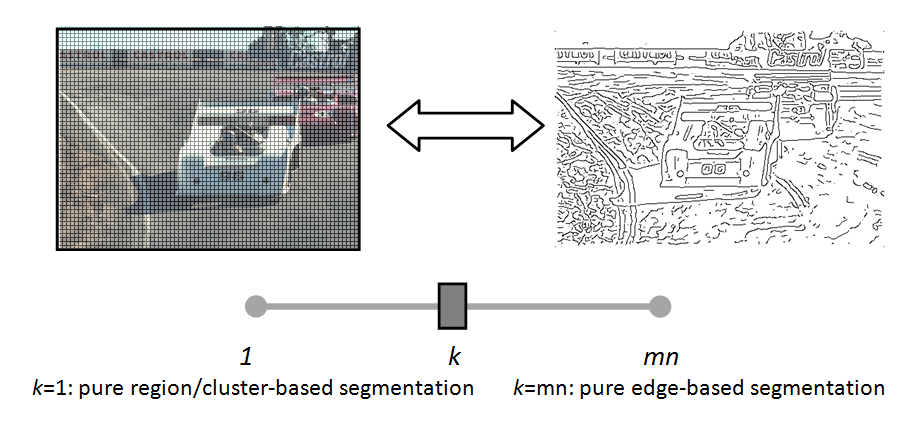}
  \end{center}
  \caption{Sliding block $k$ and entropy measurement $\ln(k)$} \label{fig:slidingblock}
\end{figure}

The value of $k$ impacts the depth of decomposition. A small value of $k$ results a deep quadtree because $\ln(k)$ is small. Small leaves do not contain too much information, which results few boundaries within the leaves. Thus $k$ is a key issue since it decides the weights of the edge- and region/clustering-based segmentation techniques used in EDHS. In other words, $k$ is a measurement that indicates the degree to which each technique is used. Figure \ref{fig:slidingblock} demonstrates that $k$ can be viewed as a sliding block ranging from $1$ to $mn$. If $k$ is close to $1$, EDHS is closer to a region/cluster-based segmentation method since few boundaries are detected in the leaves. The weight of edge-based technique increases as long as the value of $k$ becomes large. 

Suppose $k=1$, then the stopping condition $\zeta$ yields $H_k(X)\leq \ln(1)=0$. To meet this stopping condition, the decomposition process goes down to the pixel level if the neighbor pixels are inhomogeneous. Then, EDHS is a pure region/cluster-based segmentation since there is no necessary to detect the boundaries for the one color images. 

Suppose $k=mn$, then the stopping condition $\zeta$ yields $H_k(X)\leq \ln(mn)$ . By Theorem \ref{thr:entropymax}, no decomposition approach is processed since $\zeta$ holds for an $m\times n$ image by Theorem \ref{thr:entropymax}. Then, EDHS is a pure edge-based segmentation since no decomposition approach is employed. EDHS just runs an edge detector locating the boundaries to form the local regions.

For an $m\times n$ image, the possible values of $k$ range from $1$ to $mn$. Are all these integers from $1$ to $mn$ valid for $k$? The answer is no. Let us take a close look at the cases when $k=1$ and $k=2$.

Points, lines and regions are three essential elements in a two-dimensional image plane. We are looking for a value of $k$ which can efficiently recognize the lines and regions (we treat a point as noise if this point is inhomogeneous with its neighbors.) Keep in mind that the aim of decomposition is to reduce the disorder. It suggests that $k$ should be a small integer. When $k=1$, as discussed above, it forces the leaves to be one color. EDHS yields a pure region/cluster-based segmentation. Previous algorithms of this type have proved to be computationally expensive \cite{Suri2002}.

Consider $k=2$. The decomposition approach continually partitions the image until the local entropy is less than $\ln(2)$, which tends to force the leaves holding no more than two colors. Assuming a line passing through the sub-images, to recognize this line, one of the local regions of the leaves needs to be this line or part of a line. It makes the size of leaves quite small and forces the decomposition process to go down to pixel level. Under this circumstance, the time and space complexities are quite expensive. Another fatal drawback is that the small size of leaves makes the EDHS sensitive to noise. Even a pixel, which is inhomogeneous with its neighbors, could cause invalid recognition around this pixel area. 

\begin{figure}[!htb]
  \begin{center}
   \includegraphics[width=4in]{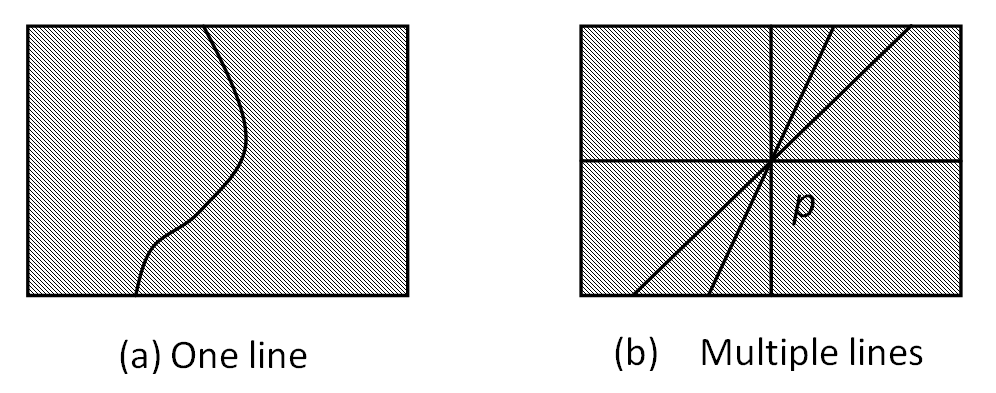}
  \end{center}
  \caption{Lines in images} \label{fig:linesinimg}
\end{figure}

$k\geq 3$ is a good choice because it can efficiently recognize the lines and regions in 2D plane. An example is shown in Figure \ref{fig:linesinimg} (a). Top-down decomposition goes down to pixel level to locate the curve if set $k<3$. But for $k\geq 3$, no decomposition is needed since the entropy of Figure \ref{fig:linesinimg} (a) must be less than $\ln(3)$ by Theorem \ref{thr:entropymax}. It suggests that EDHS is stable and reliable when $k\geq 3$.  

There is an extreme (the worst) case that we need to consider. This case is shown in Figure \ref{fig:linesinimg} (b). Multiple lines pass through one point, say $p$. Define a closed ball $B(p, \epsilon )$, where $p$ is the center and $\epsilon >0$ is the radius of the ball. No matter how small $\epsilon$ is, $B(p, \epsilon )$ contains $2N_l +1$ segments divided by $N_l$ lines. In other words, partitioning does not help reduce the number of segments around the area $B(p, \epsilon )$. Therefore, decomposing images into small pieces does not decrease entropy inside $B(p, \epsilon )$. To handle this case, we introduce `chaos' leaves.

As shown in Figure \ref{fig:linesinimg} (b), entropy does not decrease inside $B(p, \epsilon )$ along with decomposition. To solve this problem, we introduce a threshold \ref{thr:entropymax}, which is the smallest size of the leaves. If the size of partition is less than $l$, the top-down decomposition does not continue even though the desired $\zeta$ has not been reached. If this case happens, we call these leaves chaos.

\subsubsection{Approximate Image Entropy}

To calculate the entropy defined by Definition \ref{def:entropy}, we should know the probability distribution $Pr(X=x_i)=p_i$, where $X=(x_1,\ldots, x_k)$ is a set of segments of $I$. In most cases, we have no prior knowledge of the distribution of the segments for the given images. In other words, we are not able to directly compute the image entropy defined by Definition \ref{def:entropy}. Definition \ref{def:apoxentropy} gives an alternative calculation called approximate image entropy, which does not require any prior knowledge of the distribution of the segments but provides an approximate entropy value. 

\begin{mydef}[approximate entropy:]\label{def:apoxentropy} 
Given a partition $P$ of the interval $[a, b]$, $a=j_0\leq j_1 \leq\ldots\leq j_n=b$, where $a$ and $b$ are minimum and maximum of feature values of a given image $I$, then approximate entropy $H(V)$ is defined as

 \begin{equation}
 H(V)=-\sum_{i=1}^{n}p_i\log_b p_i \label{fun:apxientropy}
 \end{equation}
 
\noindent where $p$ denotes the probability mass function of the feature value set $V=\{v_1,\ldots,v_n\}$. Each $v_i$ denotes a collection of pixels whose feature values are located in $[j_{i-1},j_i]$ and $p_i=Pr(V=v_i)$, $1\leq i\leq n$. After setting the logarithm base as $e$, $H(V)$ yields

 \begin{equation}
 H(V)=-\sum_{i=1}^{n}p_i\ln p_i \label{fun:apxientropye}
 \end{equation}
 
\end{mydef}

\begin{theorem} \label{thr:aproxentropy} 
Given an $I$ and a partition $P$, $H(V)$ is less or equal then $H(X)$.
\end{theorem}

\noindent\begin{proof}

Given an $I$ and a partition $P=[j_0,j_1,\ldots,j_n]$, where $a=j_0\leq j_1\leq\ldots\leq j_n=b$, $a$ and $b$ are minimum and maximum of pixel feature values of $I$. Let $k$ is the number of segments defined in Definition \ref{def:entropy}. By the Definition of \ref{def:entropy}, $k$ must be greater or equal to $n$. There are two cases need to be considered. One is $k=n$ and another is $k>n$. 

Case I: if $k=n$, by the Definition \ref{def:entropy} and Definition \ref{def:apoxentropy}, $p(x_i)=p(v_i)$, which induces $H(X)=H(V)$.

Case II: if $k>n$, it implies that there must exist at least two segments which locate at the same partition interval. Without loss of generality, assuming $H(X)$ and $H(V)$ are defined over the partition $P$, where $H(X)$ with two segments $x_i$ and $x_j$, $1\leq i\neq j\leq k$, both feature values of $x_i$ and $x_j$ locate in the interval $[j_{h-1},j_h]$, $1\leq h\leq n$. If we can prove $H(X)>H(V)$, then the theorem can be proved by repeating following proof arbitrary times.

By Definition \ref{def:entropy} and \ref{def:apoxentropy} respectively,
\begin{align*} 
	&H(X)=-(p(x_1)\ln p(x_1)+\ldots+p(x_{h'})\ln p(x_{h'})+p(x_{h''})\ln p(x_{h''})+\ldots+p(x_n)\ln p(x_n))\\  
	&H(V)=-(p(x_1)\ln p(x_1)+\ldots+p(x_{h})\ln p(x_{h})+\ldots+p(x_n)\ln p(x_n))\\  
	&\text{where } p(x_h)=p(x_{h'})+p(x_{h''}), 0<p(x_h)<1 \text{ and } 0<p(x_{h'}), p(x_{h''})<p(x_h)\\		
	&\text{To prove } H(X)>H(V) \text{, we consider } H(X)-H(V).\\
	&H(X)-H(V)=p(x_h)\ln p(x_h)-(p(x_{h'})\ln p(x_{h'})+p(x_{h''})\ln p(x_{h''}))\\
  &\text{Let } p(x_h)=y, p(x_{h'})=x \text{, therefore } p(x_{h''})=y-x \text{, where } 0<x<y<1\\ 
  &H(X)-H(V)=y\ln y-(x\ln x+(y-x)\ln(y-x))\\
  &=\ln y^y-(\ln x^x+\ln (y-x)^{(y-x)})\\
  &=\ln y^y-\ln x^x(y-x)^{(y-x)}\\
  &\text{Let } f(x,y)=y^y-x^x(y-x)^{(y-x)}\\
  &\text{If we can prove } f(x,y)>0 \text{ for all } 0<x<y<1 \\
	&\text{then } H(X)-H(V)=\ln y^y-\ln x^x(y-x)^(y-x)>0\\
	&\text{ since } \ln(x) \text{ is an increasing function.}\\
	&f(x,y)=y^y-x^x(y-x)^{(y-x)}\\
	&=y^y-x^x\frac{(y-x)^y}{(y-x)^x}\\
	&=y^y-(\frac{x}{y-x})^x(y-x)^y\\
	&=y^y(1-(\frac{x}{y-x})^x(\frac{y-x}{y})^y) \\
	&\text{Let } \alpha=\frac{x}{y-x} \text{, then} f(x,y)=y^y(1-\alpha^x(\frac{1}{1+\alpha})^y)=y^y(1-\frac{\alpha^x}{(1+\alpha)^y})\\
  &\text{Notice that } \alpha^x<(1+\alpha)^y \text{ since } \alpha \text{ is a positive number and } 0<x<y<1\\
	&\text{Then, }  1-\frac{\alpha^x}{(1+\alpha)^y} \text{ is positive. It implies } f(x,y)>0 \text{ because } y^y>0	
\end{align*} 

\end{proof}

Assume an $I$ and $k$-CR is true. It implies that $H_k(X)\leq \ln(k)$ by Theorem \ref{thr:entropymax}. By Theorem \ref{thr:aproxentropy}, $H(V)$ must be equal or less than $H_k(X)$. Therefore, It induces the a logical chain, truth of $k$-CR $\Rightarrow H_k(X)\leq \ln(k) \Rightarrow H(V)\leq \ln(k)$. Both $H_k(X)\leq\ln(k)$ and $H(V)\leq\ln(k)$ are necessary but not sufficient conditions for the truth of $k$-CR. 

\subsubsection{Noise Segments}

The term noise of an image usually refers to unwanted color or texture information. We do not count a small drop of ink on an A4 paper as a valid segment. In most circumstances, it would be considered as noise. How to distinguish those valid and invalid segments is an important issue. 

\begin{mydef}[Dominant and Noise segments:]\label{def:dnsegment} 
Let $p$ denote the probability mass function of segment set $X=\{x_1,\ldots,x_k\}$ in $I$. Given a threshold $t_{noise}\in[0, 1]$, $x_i$ is a noise segment if $Pr(X=x_i)=p_i\leq t_{noise}$ and $\sum_{p_i\leq t_{noise}} p_i$ is greatly less than $\sum_{p_i>t_{noise}} p_i$ , $1\leq i \leq k$. Other segments are called dominant segments.
\end{mydef}

If the segments are small enough and the total area of those segments occupies a small portion of a given image, we call those segments noise segments. The first requirement of noise segment is understandable because the noise segments should be small. The reason of defining the second condition is to avoid the cases that the images are totally composed by small pieces. 

The value of $k$ of K-CR in Definition \ref{def:kcr} refers the number of dominant segments. By Theorem \ref{thr:entropymax} and \ref{thr:aproxentropy}, the supremum of $H(V)$ for this given image is no longer $\ln(k)$. The noise supremum of $H(V)$ should be slightly larger than $ln(k)$. Assuming the noise redundancy be $\epsilon$, then redundancy stopping condition, $\zeta _r$, yields $H(V)\leq \ln(k)+\epsilon$.

Consider dividing segments into two groups, noise and dominant segments. By Definition \ref{def:apoxentropy}, $H(V)$ yields as follows:
\[
H(V)=-\sum_{x_i\in X} p(x_i)\ln p(x_i)=-\sum_{p(x_i)>t_{noise}}p(x_i)\ln p(x_i)-\sum_{p(x_i)\leq t_{noise}}p(x_i)\ln p(x_i)
\]
Given an image $I$, let $a$ be the total portion of dominant segments. Then $a=\sum_{p(x_i)>t_{noise}} p(x_i)$ and the rest area, $1-a=\sum_{p(x_i)\leq t_{noise}} p(x_i)$, is the portion of noise segments. By Definition \ref{def:dnsegment}, $a>>(1-a)$. Let $k$ and $k'$ be the number of dominant and noise segments respectively. After applying Theorem 2.6.4 \cite{Cover1991}, we get the noise supermum of $H(V)$ as follows.
\[
H(V)=-\sum_{p(x_i)>t_{noise}}p(x_i)\ln p(x_i)-\sum_{p(x_i)\leq t_{noise}}p(x_i)\ln p(x_i)\leq -(a\ln(\frac{a}{k})+(1-a)\ln(\frac{1-a}{k'}))
\]
The noise redundancy $\epsilon=-(a\ln(\frac{a}{k})+(1-a)\ln(\frac{1-a}{k'}))-\ln(k)$. The redundancy stopping condition, $\zeta _r$, yields $H(V)\leq -(a\ln(\frac{a}{k})+(1-a)\ln(\frac{1-a}{k'}))$.

Following gives an example to compute the noise redundancy. Suppose a 3-CR application $(k=3)$, setting $a=0.98$ and $k'=3$. The redundancy stopping condition for 3-CR yields $-(a\ln(\frac{a}{k})+(1-a)\ln(\frac{1-a}{k'}))=1.1968$, which is slight greatly than $ln(k)=1.0986$. Noise redundancy $\epsilon=-(a\ln(\frac{a}{k})+(1-a)\ln(\frac{1-a}{k'}))-\ln(k)=0.0981$.

\begin{figure}[!htb]
  \begin{center}
   \includegraphics[width=5in]{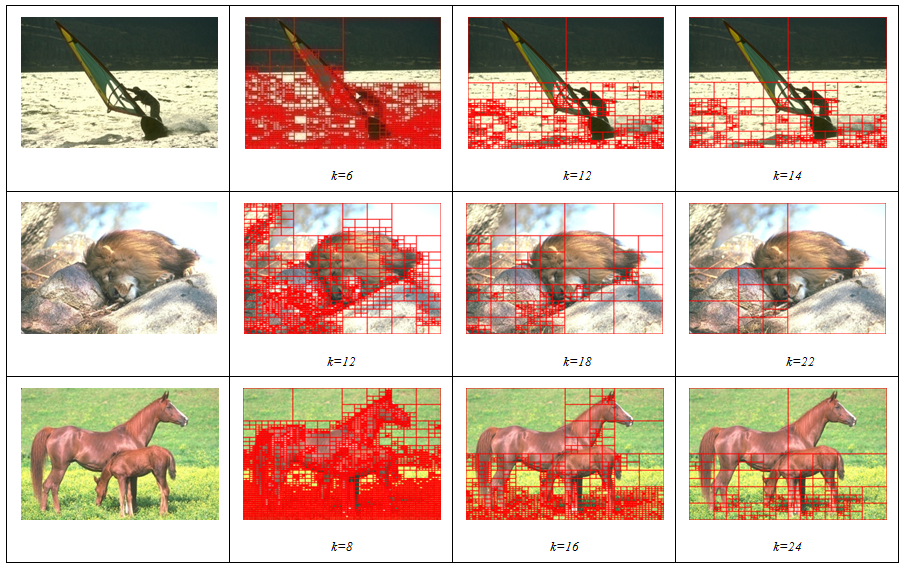}
  \end{center}
  \caption{Quarter decomposition by different stopping conditions} \label{fig:exampledecomposition}
\end{figure}

We summarize the top-down decomposition by Algorithm \ref{algo:topdowndecomposition} and demonstrate some examples of the top-down decomposition in Figure \ref{fig:exampledecomposition} by varying different $k$ values, where $l=3$, $a=0.998$ and $k'=3$.

\incmargin{1em}
\restylealgo{boxed}\linesnumbered
\begin{algorithm}[H]
\SetKwFunction{FindCompress}{FindCompress}
\SetKwInOut{Input}{input}
\SetKwInOut{Output}{output}
\caption{Topdowndecomposition}
	\Input{$I$: An image}
	\Output{$T_{qud}$: A decomposition quadtree}
\BlankLine
\eIf{size of $I$ $<l$}{ 
  // current $I$ is chaos	\\
	Create a chaos leaf for $I$ and generate a feature descriptor for $I$.\\ 
}{
	\eIf{$H(V)> -(a\ln(\frac{a}{k})+(1-a)\ln(\frac{1-a}{k'}))$}{
	  Partition $I$ into four partitions: $quad1$, $quad2$, $quad3$, $quad4$;\\
	  Append $quad1$, $quad2$, $quad3$ and $quad4$ as children of $I$ in the $T_{qud}$;\\
	  Topdowndecomposition($quad1$);\\
	  Topdowndecomposition($quad2$);\\
	  Topdowndecomposition($quad3$);\\
	  Topdowndecomposition($quad4$);\\
	}{
	    Locate the local regions by detecting the boundaries within $I$; \\
	    Create a non-chaos leaf and generate a feature descriptor for each local region; \\
	}
}
\label{algo:topdowndecomposition}
\end{algorithm}
\decmargin{1em}

\subsection{Bottom-up Composition}

Bottom-up composition stands at a kernel position of VHBS since this process combines the local regions at the leaves of the quadtree to form the initial segment set $S=(s_0,\ldots,s_n)$. It also calculates the probabilities of the boundaries between these initial segments. At the same time, bottom-up composition process generates the feature descriptors for each initial segment by combining the local region feature descriptors. 

The probabilities of the boundaries between these initial segments are computed by two filters, which are designed based on the two visual hint rules (i) and (ii) separately. The first one called scale filter, $f_1$, abides by rule (i). The probabilities are measured by the length of the boundaries. Longer boundaries result higher probabilities. The second one called similarity filter, $f_2$, abides by rule (ii). The probabilities of the boundaries are measured by the differences of two adjacent regions. Larger different features of two adjacent regions result higher probability boundaries. The finial weights of the boundaries are the trade-off of two filters by taking the products of these two filters. If the probability of the boundary between two local regions is zero, these two local regions are combined together. 

\subsubsection{Scale Filter}\label{sec:scalefilter}

Scale filter is defined based on the visual hint (i): the global scale boundaries tend to be the real boundaries of the objects. It suggests that these boundaries caused by the local texture are not likely to be the boundaries of our interesting objects because the objects with large size are more likely to be our interesting objects. To measure the relative length of each boundary, we use the sizes of the decomposition partitions in which the objects are fully located. These local scale boundaries are not likely to extend to a number of partitions since the length of these boundaries are short. By this observation, we define the scale filter $f_1$ based on the sizes of the partitions.  

\begin{figure}[!htb]
  \begin{center}
   \includegraphics[width=4in]{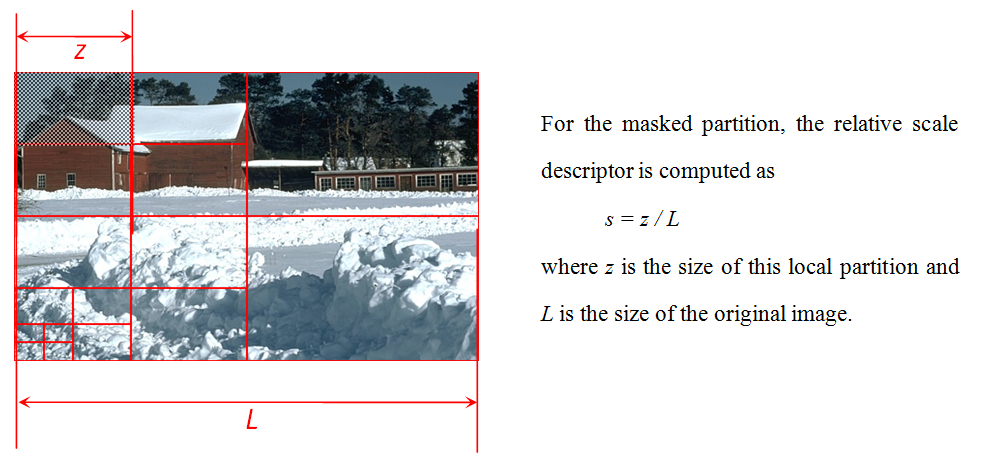}
  \end{center}
  \caption{The relative scale descriptor} \label{fig:scaledescriptor}
\end{figure}

Scale filter $f_1$ is a function which calculates the confidence of the boundary based on the scale observations. The input parameter of $f_1$ is the relative scale descriptor $s$, which is ratio between the sizes of the local partitions and the original images. The relative scale descriptor $s$ is the measurement of the relative scale of the boundaries. We assume the sizes of the images are the length of the longer sides. An example of calculating scale descriptor $s$ is shown in Figure \ref{fig:scaledescriptor}. The boundaries inside the marked partition have the relative scale descriptor $s$, which is defined as:
 \begin{equation}
 s=z/L \label{fun:scaledescriptor}
 \end{equation}
where $z$ is the size of a partition and $L$ is the size of a original image.

\begin{figure}[!htb]
  \begin{center}
   \includegraphics[width=4in]{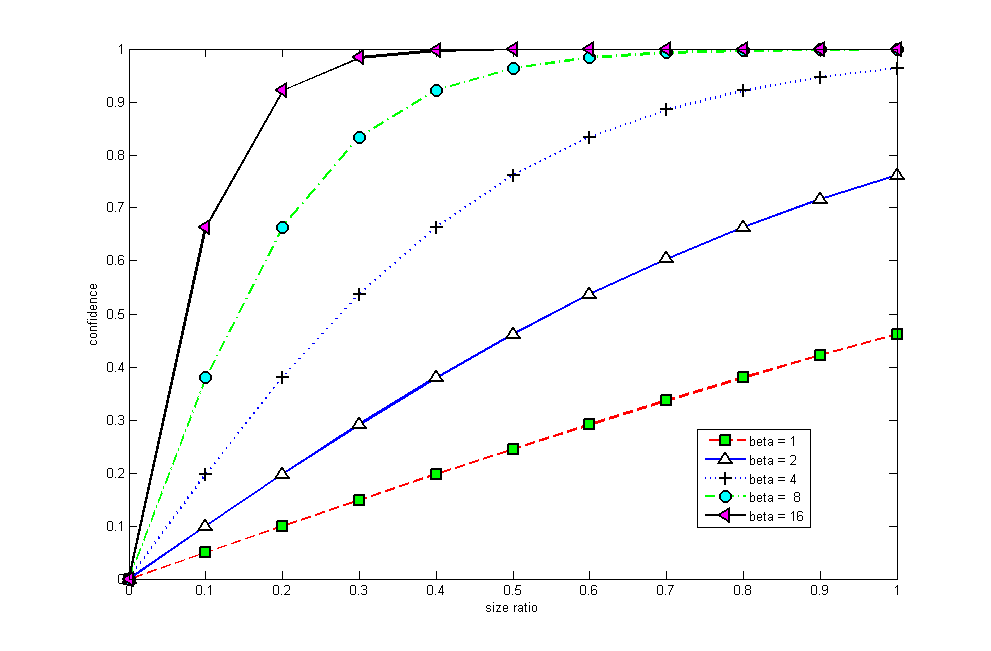}
  \end{center}
  \caption{Scale filter $f_1(s)$ with different scale damping coefficients.} \label{fig:scalefilter}
\end{figure}

Based on visual hint (i), $f_1(s)$ must be a strictly increasing function on domain $[0, 1]$ and the range of $f_1(s)$ locates in the interval $[0, 1]$. If s is small, it suggests that the confidence of the boundary should be small since the boundaries are just located in a small area. If $s$ is close to one, it suggests that the boundaries have high confidence. The gradient of $f_1(s)$ is decreasing. This is because human perception is not linearly dependent on the relative scale descriptor. For the same difference, human perception is more sensitive when both of them are short rather than both of them are long. Therefore we define $f_1(s)$ as follows:
 \begin{equation}
 f_1(s)=\frac{2}{1+e^{-\beta_1 s}}-1 \label{fun:scalefilter}
 \end{equation}
where $\beta_1$ is the scale damping coefficient and $s$ is the relative scale descriptor. Figure \ref{fig:scalefilter} gives the $f_1(s)$ with different scale damping coefficients. 

\subsubsection{Similarity Filter}

Compared these regions with similar colors or textures, human perception is more impressed by regions with quite different features. Visual hint (ii) suggests that two adjacent regions with different colors or textures tend to produce the high confidence of boundaries between them. Based on this observation, we defined a similarity filter $f_2$ to filter out the boundary signals which pass through similar regions.

\begin{figure}[!htb]
  \begin{center}
   \includegraphics[width=4in]{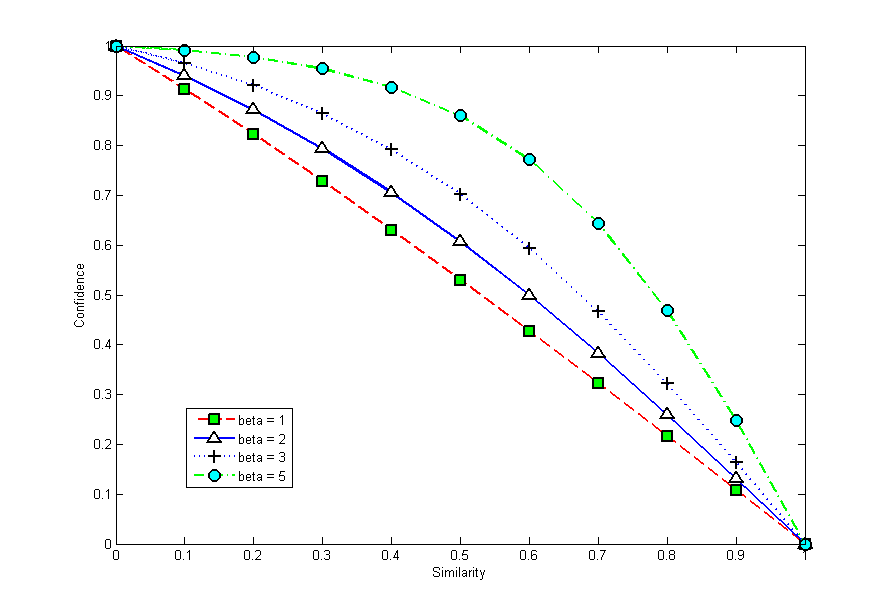}
  \end{center}
  \caption{Similarity filter $f_2(s)$ with different similarity damping coefficients.} \label{fig:similarityfilter1}
\end{figure}

Similarity filter $f_2$ is a function which calculates the confidence of the boundaries based on the similarity measurements between two adjacent regions. Examples of similarity measurements are Dice, Jaccard, Cosine, Overlap. The similarity measurement is a real number in interval $[0, 1]$. The higher the value, the more similar two regions are. $Similarity=1$ implies that they are absolutely the same and zero means they are totally different. Based on the visual hint (ii), small similarity measurement should result in high confidence boundary. Let $x$ denote the similarity measurement of two adjacent regions. Then $f_2(x)$ is a strictly decreasing function over domain $[0, 1]$ and the range of $f_2(x)$ is inside $[0, 1]$. Human perception is not linear relationship with the similarity measures. Human perception is sensitive to the regions when these regions have obvious different colors or textures. For example, similarity measure $0.1$ and $0.2$ are not a big difference for human visual because both of them are obviously different. This fact suggests that the gradient of $f_2(x)$ is decreasing over the domain $[0, 1]$. Then we define $f_2(x)$ as follows to satisfy the requirements above.
 \begin{equation}
 f_2(x)=(\frac{2}{1+e^{-\beta_2 (1-x)}}-1)(\frac{1+e^{-\beta_2}}{1-e^{-\beta_2}}) \label{fun:similarityfilter1}
 \end{equation}
where $\beta_2$ is the similarity damping coefficient and $x$ is the similarity measurement between two adjacent regions defined as $x=similar(fd_1, fd_2)\in[0, 1]$. $fd_1$ and $fd_2$ are feature descriptors of the local regions. Figure \ref{fig:similarityfilter1} gives the curves of $f_2(x)$ with respect to different similarity damping coefficients. 

\begin{figure}[!htb]
  \begin{center}
   \includegraphics[width=4in]{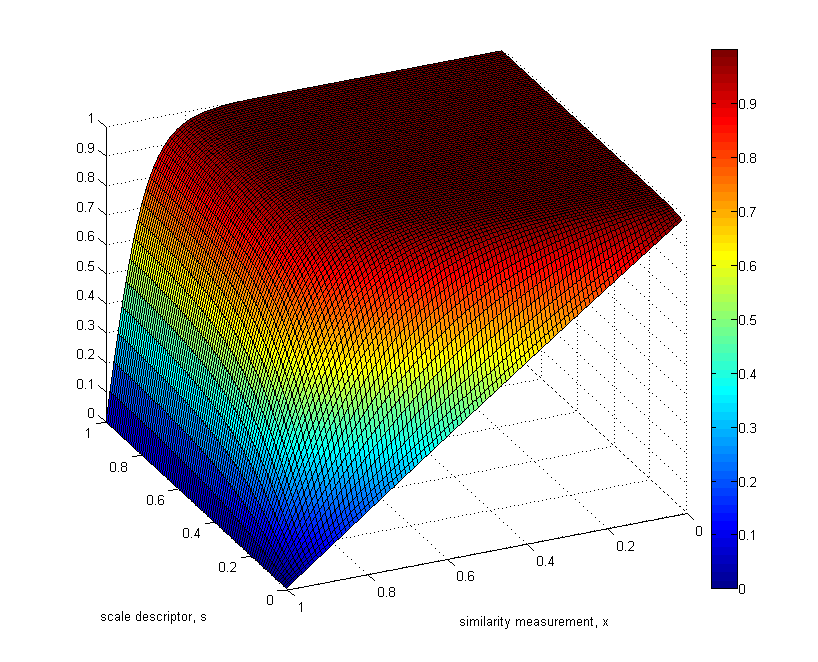}
  \end{center}
  \caption{Similarity filter $f_2(x,s)$ with $\beta_2=10$ and $\alpha=20$.} \label{fig:similarityfilter}
\end{figure}

We implemented VHBS by using $f_2(x)$ given by equation \ref{fun:scalefilter}. We found that the boundary signals are over-damped by the similarity filter $f_2(x)$. The algorithm assigns low confidence values for boundaries with global scales that preserved similar feature descriptors. Human visual is also sensitive to these sorts of boundaries. To avoid these cases, we redesign the similarity filter by considering the relative scale descriptor as well. Similarity filter is redefined as $f_2(x,s)$, which outputs high confidence weights when either parameter $s$ or $x$ is close to one. We also introduce a threshold similarity $t$. If the similarity measurement is higher than this threshold, algorithm sets the confidence as $0$, which means that there are no boundaries between these two regions if human visual cannot tell the difference of these two adjacent regions.
 \begin{equation}
f_2(x,s) = \left\{ 
\begin{array}{l l}
  (\frac{2}{1+e^{-y(1-x)}}-1)(\frac{1+e^{-y}}{1-e^{-y}})  & \quad \text{if $x<t$}\\
  \text{      where } y=\alpha (\frac{2}{1+e^{-\beta_2 s}}-1)(\frac{1+e^{-\beta_2}}{1-e^{-\beta_2}}) \\
  0 & \quad \text{if $x\geq t$}\\
\end{array} \right.
\label{fun:similarityfilter}
\end{equation}

where $\beta_2$ is the similarity damping coefficient, $\alpha$ is amplitude modulation and $x$ is the similarity measurement between two feature descriptors. Figure \ref{fig:similarityfilter} demonstrates $f_2(x,s)$ when $\beta_2=10$ and $\alpha=20$.

\subsubsection{Partition Combination}

The bottom-up composition starts from the very bottom leaves. Composition process iteratively combines partitions from the next lower layer and this process continues until reaching the root of the quadtree.

\begin{figure}[!htb]
  \begin{center}
   \includegraphics[width=4in]{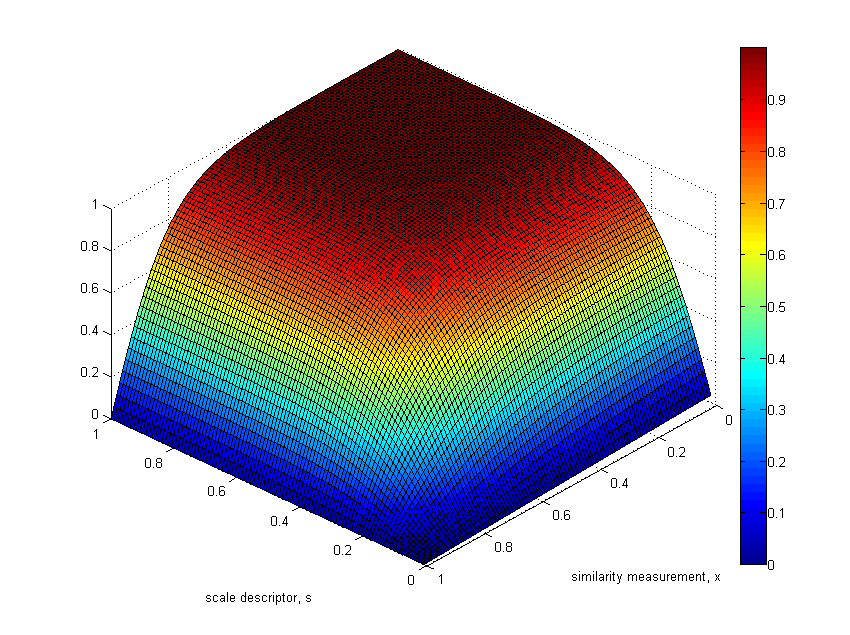}
  \end{center}
  \caption{$f_1(s)\cdot f_2(x,s)$ with $\beta_1=8$, $\beta_2=10$ and $\alpha=20$.} \label{fig:f1f2}
\end{figure}

For the leaves of the quadtree, each boundary is marked by a confidence value, $cnf$, which is given by formula $f_1(s)\cdot f_2(x, s)$, where $f_1(s)$ and $f_2(x,s)$ are scale and similarity filters defined by equations \ref{fun:scalefilter} and \ref{fun:similarityfilter} respectively. The relative scale descriptor, $s$, is computed by equation \ref{fun:scaledescriptor} and $x$ is the similarity of two adjacent local segments. Figure \ref{fig:f1f2} demonstrates the function $cnf=f_1(s)\cdot f_2(x,s)$ where with $\beta_1=8$, $\beta_2=10$ and $\alpha=20$.

\begin{figure}[!htb]
  \begin{center}
   \includegraphics[width=5in]{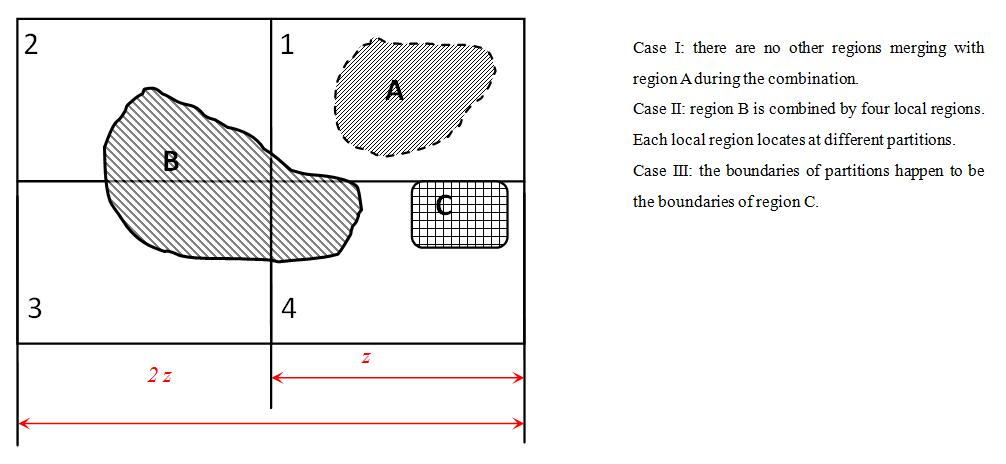}
  \end{center}
  \caption{Three cases of partition combination.} \label{fig:casescombination}
\end{figure}

For these non-chaos leaves, the contours which form the closed areas and the borders of the leaves form the boundaries of local regions. For these chaos leaves, the boundaries only refer to the leaf borders. Each running time of bottom-up combination, four leaves are combined together to the next lower layer. Figure \ref{fig:casescombination} shows that there are several possible cases during combining four leaves together.


\begin{enumerate}
 \item No interconnection happens during the combination;
 \item A new segment is formed by connecting several local regions which locate in different leaves;
 \item The boundaries of leaves happen to be the boundaries of segments.
\end{enumerate}

For the case (i) such as region A shown in Figure \ref{fig:casescombination}, $cnf$ is calculated when it is in leaf $1$ and there is no necessary to recalculate $cnf$ during the combination. But for the case (ii) such as region B, region B is connected by four local regions. Each $cnf$ is calculated separately. But the $cnf$  region B needs to be recalculated after four local regions combined together since the new combined segment is located in a large partition and the relative scale descriptor is increased. Besides, a feature descriptor for region B is generated based on feature descriptors for each local region. The third case is that the boundaries of partitions happen to be the boundaries of the regions. One example is region C shown in Figure \ref{fig:casescombination}. During the combination process, algorithm also calculates $cnf$ of the leaf boundaries. After combination hits the root of the quadtree, the process generates the initial segment set $S=(s_0,\ldots,s_n)$, the boundary confidence set $C_b=\{c_{i,j}\}$, where $c_{i,j}$ indicates the boundary probability between segments $s_i$ and $s_j$, $0\leq i\neq j\leq n$ and the feature descriptor set $FD=(fd_0,\ldots,fd_n)$. Follows is the Algorithm \ref{algo:bottomupcomposition} of iterative bottom-up composition.

\incmargin{1em}
\restylealgo{boxed}\linesnumbered
\begin{algorithm}[H]
\SetKwFunction{FindCompress}{FindCompress}
\SetKwInOut{Input}{input}
\SetKwInOut{Output}{output}
\caption{Bottomupcomposition}
	\Input{$T_{qud}$: A decomposition quadtree \\ $id$: root of $T_{qud}$}
	\Output{$S=(s_0,\ldots,s_n)$: the initial segment set \\ $C_b=\{c_{i,j}\}$: the boundary confidence set\\ $FD=(fd_0,\ldots,fd_n)$: the feature descriptor set}
\BlankLine
node$\leftarrow$read a node of $T_{qud}$ by $id$; \\
\eIf{node has children}{ 
  Read the four children of node as, $id1$,$id2$,$id3$,$id4$; \\
  quad1$\leftarrow$Bottomupcomposition($id1$);\\
  quad2$\leftarrow$Bottomupcomposition($id2$);\\
  quad3$\leftarrow$Bottomupcomposition($id3$);\\
  quad4$\leftarrow$Bottomupcomposition($id4$);\\
  //Combine partitions. Recompute $cnf=f_1(s)\cdot f_2(x,s)$ if needed \\
	img$\leftarrow$combine(quad1,quad2,quad3,quad4); \\
}{
	//This node is a leaf of $T_{qud}$\\
	img$\leftarrow$computelocalcnf(node); //$cnf=f_1(s)\cdot f_2(x,s)$\\
	return img;
}
\label{algo:bottomupcomposition}
\end{algorithm}
\decmargin{1em}

\section{Hierarchical Probability Segmentation}

Hierarchical segmentation is a widely used technique for image segmentation. Regular hierarchical segmentation is modeled in layer-built structures. Compared with the regular hierarchical structures, Hierarchical Probability Segmentation, HPS, presents the hierarchical segmentation by a Probability Binary Tree (PBT,) where the links are weighted by the confidence values, $cnf\in[0, 1]$. The root represents an image. Nodes represent segments and the children of a node are the sub-segments of this node. Initial segments $S=(s_0,\ldots,s_n)$ compose the leaves of PBT. Since PBT is generated in greedy manner, higher level nodes always have higher probabilities than the lower level nodes. One can visualize the PBT in arbitrary number of segments. Of course, this number is less than the number of the initial segments.

\subsection{Probability Binary Tree (PBT)}

\begin{mydef}[Probability Binary Tree (PBT):]\label{def:PBT} 
Let $n_0$ denote the root of a PBT and $n_0$ represents the original images $I$. Nodes of PBT denote the segments and links represent the relationship of inclusion. Assume nodes $n_i$, $n_{i1}$, $n_{i2}$ and links $l_{i1}$, $l_{i2}$, where $n_{i1}$ and $n_{i2}$ are children of $n_i$ linked by $l_{i1}$ and $l_{i2}$ respectively. $n_i$, $n_{i1}$, $n_{i2}$, $l_{i1}$, $l_{i2}$ preserve the following properties:
\begin{enumerate}
 \item Let $DS=\{(n_{i1}^1,n_{i2}^1),(n_{i1}^2,n_{i2}^2),\ldots\}$ denote the set of all the possible pairs of sub-segments of $n_i$. Function $g(n_{i1}^j,n_{i2}^j)$ gives the $cnf$ of segments $n_{i1}^j$ and $n_{i2}^j$. Assume $(n_{i1},n_{i2})=\arg \max g(DS)$ and $l_{i1}$. Therefore $l_{i2}$ are weighted by $g(n_{i1}, n_{i2})$;
 \item for any element of $DS$, $(n_{i1}^j,n_{i2}^j)$, $n_i=n_{i1}^j \cup n_{i2}^j$ and $\phi=n_{i1}^j \cap n_{i2}^j$.
\end{enumerate}
 
\end{mydef}

Definition \ref{def:PBT} recursively gives the definition of PBT, which has the following properties:
\begin{enumerate}
 \item Every PBT node (except root) is contained in exactly one parent node;
 \item every PBT node (except the leaves) is spanned by two child nodes;
 \item a number of pairs of nodes span $n_i$. These pairs are candidates to be the children of $n_i$ and each pair is labeled with the probability of these two nodes, $\{cnf_0=g(<n_{i1}^0, n_{i2}^0>), cnf_1=g(<n_{i1}^1, n_{i2}^1>),\ldots\}$. PBT chooses the pairs $<n_{i1}^j, n_{i2}^j>$, which have the highest $cnf_j$ to span the node $n_i$. The links $l_{i1}$ and $l_{i2}$ are weighted by $cnf_j$;
 \item assume a node $n_i$ with a link $l_i$ pointed in from its parent and the links $l_{i1}$, $l_{i2}$ pointing out to its children. Weights of $l_{i1}$, $l_{i2}$ must be no larger than the weight of $l_i$;
 \item if two nodes (segments) overlap, one of them must be a child of the other.
\end{enumerate}

\begin{figure}[!htb]
  \begin{center}
   \includegraphics[width=5in]{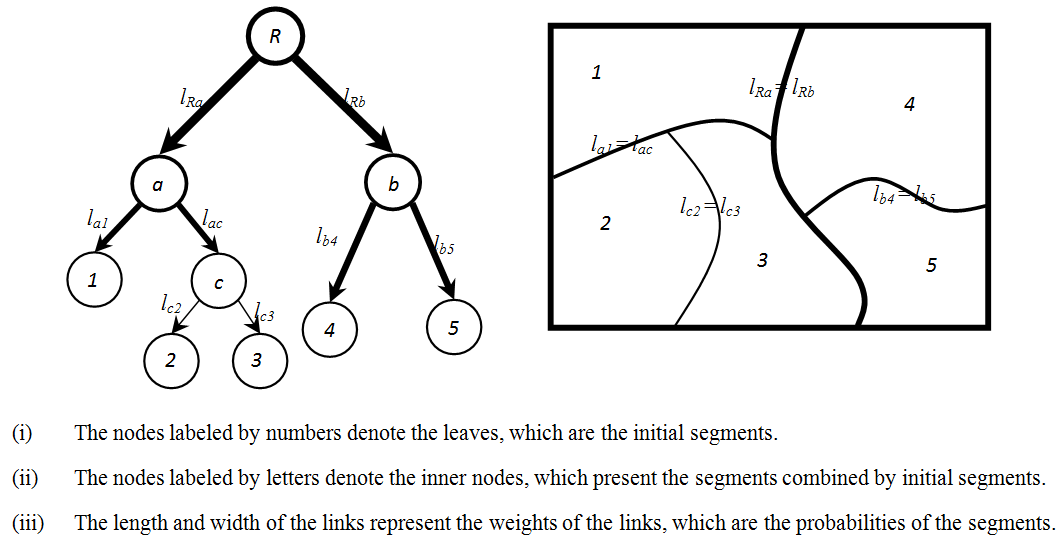}
  \end{center}
  \caption{Probability Binary Tree.} \label{fig:PBT}
\end{figure}

By presenting the segmentation in PBT, the images are recursively partitioned in two segments with the highest probability among all the possible pair of segments. Figure \ref{fig:PBT} gives an example of the PBT.

Let $C_b=\{c_{i,j}\}$ denote the set of boundary probabilities. $c_{i,j}\in C_b$ represents the boundary probability between initial segments $s_i$ and $s_j$, where $c_{i,j}\in[0, 1]$, $0\leq i\neq j\leq n$. HPS constructs PBT in bottom-up manner, which means leaves are first created and the root is the last node created. To generate a PBT in greedy schema, $C_b$ is sorted in ascending order. Let $T_{pbt}$ be a PBT and $C'_b$ be the ascending order sequence of $C_b$. Algorithm \ref{algo:PBT} describes to generate a $T_{pbt}$. 

\incmargin{1em}
\restylealgo{boxed}\linesnumbered
\begin{algorithm}[H]
\SetKwFunction{FindCompress}{FindCompress}
\SetKwInOut{Input}{input}
\SetKwInOut{Output}{output}
\caption{Generating a probability binary tree.}
	\Input{$C'_b$: sorted $C_b$ in ascending order}
	\Output{$T_{pbt}$: a probability binary tree, PBT}
\BlankLine
\While{$C'_b$ is not empty} {
  $c_{ij} \leftarrow$ read the first element of $C'_b$ and remove it from $C'_b$;\\
  $i\leftarrow$ read the index of segment $s_i$ from $c_{ij}$;\\
  $j\leftarrow$ read the index of segment $s_j$ from $c_{ij}$;\\
	\eIf{$s_i$ is not exist in $T_{pbt}$}{
		Create a node $s_i$ in $T_{pbt}$;
	}{
		$s_i\leftarrow$ read the $s_i$ from $T_{pbt}$;
	} 
	\eIf{$s_j$ is not exist in $T_{pbt}$}{
		Create a node $s_j$ in $T_{pbt}$;
	}{
		$s_j\leftarrow$ read the $s_j$ from $T_{pbt}$;
	} 	
	Create a new node $n_{new}$, which is the parent of $s_i$ and $s_j$;\\
	Create the links from $n_{new}$ to $s_i$ and from $n_{new}$ to $s_i$ weighted by $c_{ij}$;\\
	Replace index $i$ and $j$ in current $C'_b$ by the index of $n_{new}$ and remove the duplicate elements in $C'_b$;
}
\label{algo:PBT}
\end{algorithm}
\decmargin{1em}

\subsection{Visualization of the Segmentation}

Generally, there are two ways to visualize the segments. One is threshold-based visualization and another is number-based visualization. As discussed in previous section, the root of the PBT represents the original images. For the other nodes, the more shallow the positions are, the more coarse-gradient the segments are. For example, visualizing image in segments $a$ and $b$ shown in Figure \ref{fig:PBT} is combining initial segments $1$, $2$ and $3$ together to form segment $a$ and combining initial segments $4$ and $5$ together to form segment $b$. 

Suppose a threshold, $t_{visual}\in[0, 1]$, is selected for visualization. By the properties of the PBT, the weights of the links are the probabilities of the segments. The weighs are decreasing as long as the depths are increasing. Given a $t_{visual}$, threshold-based visualization only displays the segments whose link weights are greater than the given $t_{visual}$. 

The number-based visualization displays a certain number of segments. Let $n_{visual}$ denote the number of visualization segments. The implementation of number-based visualization is trivial. Algorithm sorts the nodes in descending order with respect to the link weights and picks the first $n_{visual}$ number of nodes to display. 

\section{Algorithm Complexity Analysis}

Since the algorithm is divided into two stages, EDHS and HPS, we discuss the computational complexity of them separately. 

\subsection{EDHS Computational Complexity}

Assume the depth of the quadtree generated by top-down decomposition is $d$ and the depth of the root is zero. The maximum $d$ is $(\min(\lfloor\log_2 n\rfloor, \lfloor\log_2 m\rfloor)-\lceil\log_2 l\rceil)$, where $n\times m$ is the size of original images and $l$ is the chaos threshold. Depending on the different images and the chosen stopping condition $\zeta$, decomposition process generates an unbalance quadtree with depth of $d$. To analysis the complexity of the decomposition, we assume the worst cases that the images are fully decomposed. It implies that depth of all the leaves is $d=(\min(\lfloor\log_2 n\rfloor, \lfloor\log_2 m\rfloor)-\lceil\log_2 l\rceil)$. At the $i$th level of the quadtree, there are $4^i$ numbers of nodes and the size of each node is $\frac{mn}{4^i}$. Then the running time of computing the stopping condition $\zeta$ of the ith depth is $4^i\frac{mn}{4^i} = mn$ and the total running time of decomposition is $dmn$. Commonly, the time complexity of an edge detector is $O(mn)$ such as Canny Edge Detection \cite{Canny1986}. Plus the time complexity of generating the feature descriptors $O(mn)$. The running time of top-down decomposition is $(d+2)mn$, which gives the time complexity of top-down decomposition $O(mn)$. 

The combination process starts from the leaves to calculate the boundary confidence by $cnf=f_1(s)\cdot f_2(x,s)$, which gives the running time $\frac{m}{2^d}\frac{n}{2^d}$ for each leaf (leaf size is $\frac{m}{2^d}\frac{n}{2^d}$) and total running time is $4^d\frac{mn}{4^d} = mn$ since there are $4^d$ numbers of leaves totally. At the $i$th level of the quadtree, the composition process combines the four quadrants into one, which gives the running time $m+n+\frac{m}{2^i}\frac{n}{2^i}$ and total running time of the $i$th depth is $4^i(m+n+\frac{m}{2^i}\frac{n}{2^i})=mn+4^i(m+n)$. Then the total running time of bottom-up composition is $(d+1)mn+\frac{4^d-1}{3}(m+n)$. It can be proved when $d$ is large enough, term of $\frac{(4^d-1)(m+n)}{3}$ dominates the running time. It gives the time complexity $O(4^d(m+n))$. Then the time complexity of the EDHS is $O(4^d(m+n))$, where $d$ is the depth of the quadtree and mn is the size of the input images.

\subsection{HPS Computational Complexity}

To make the analysis simple, we assume maximum $d$ is $log_2 n$ (under the worst situation.) It suggests that the maximum number leaves is $4^{log_2 n}=n^2$. This implies the worst running complexity case is that one pixel is one local segment. At pixel level, PHS considers four pixel neighbors as adjacent local segments. They are left, right, upper and below pixels. Then the maximum size of $C_b$ is $2mn$. PHS first sorts $C_b$ to $C'_b$, which gives the time complexity $O(mn\log(mn))$. Considering the algorithm to generate the $T_{pbt}$, the total running time is $\sum_{i=1}^{2mn} (2mn-i)=2(mn)^2-mn$. This gives time complexity $O((mn)^2)$. Compare with the complexity of EDHS, generating $T_{pbt}$ is more expensive part. In real applications, the number of elements in $C_b$ is far less than $2mn$ since the algorithm does not go down to pixel level by setting the size threshold of the partitions. The running time also highly depends on the input images because the number of local segments function depends on the complexity of the image itself. For the test running experiments of image set \cite{Martin2001}, the average number of elements of $C_b$ is around $3000$ or less. Then the practical time complexity of algorithm is far less than $O((mn)^2)$.

\section{Experimental Results}

In this section, we demonstrate experiments of VHBS by tuning different parameters and evaluate the results by comparing with outputs of Normalized Cut \cite{Shi1997} and KMST \cite{Felzenszwalb1998,Felzenszwalb2004} based on the same test set of Berkeley Segmentation dataset (BSDB) \cite{Martin2001}.

\subsection{Weighted Boundary Evaluations}

As discussed in previous section, the visualization of boundary probability set $C_b$ results in the weighted boundaries of the initial segments $S=(s_0,\ldots,s_n)$. Some examples are demonstrated in Figure 10. We use the benchmark provided by BSDB, which uses Precision-Recall curves \cite{Rijsbergen1979} to evaluate the boundary detection. Precision is defined as the number of true positives over the number of elements retrieved by the detection, and Recall is defined as the number of elements retrieved by detection divided by the total number of existing relevant elements. Precision measure can be viewed as the correctness and Recall measure can be viewed as the completeness. 

Since the purpose of our algorithm is to locate the segments, algorithm only marks the contours that form the boundaries of the regions. Incontinuities or unclosed contours are ignored. These ignored contours include three portions. One is the contours within the chaos leaves. The second is the contours in non-chaos leaves that are Incontinuities or unclosed. These boundaries between two regions with similarity measure higher than the threshold $t$ (in equation \ref{fun:similarityfilter}) are also ignored. These suggest that Precision-Recall curves mainly located at the left side of the PR graphs.

\begin{figure}[!htb]
  \begin{center}
   \includegraphics[width=4in]{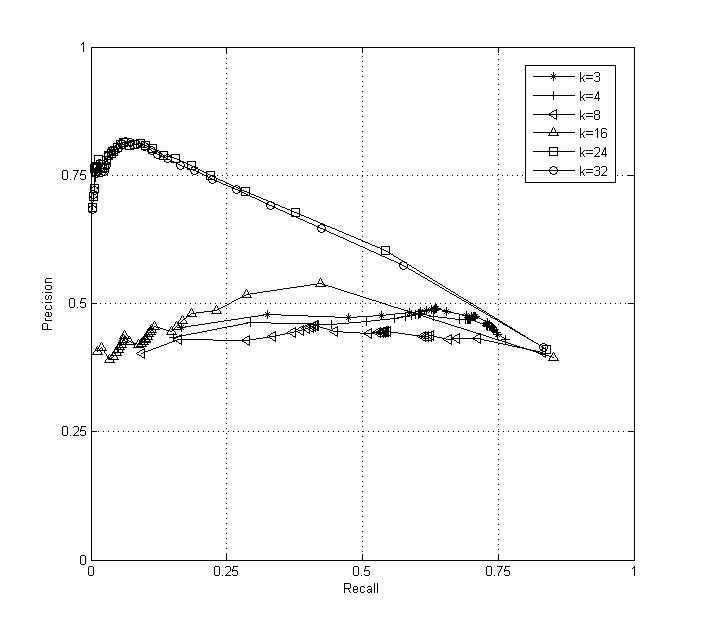}
  \end{center}
  \caption{Boundary evaluation based on $k$.} \label{fig:boundaryevlonk}
\end{figure}

The first set of experiments is based on the observation of different values of $k$. We discussed the value of $k$ in section \ref{sec:valueofk}. Recall one of the purposes of decomposition is to reduce the information contained in leaves. From this point of view, small integer is preferred for $k$ since small values of $k$ result more strict stopping condition. On the other hand, we need to avoid too many chaos leaves since chaos means that the algorithm has failed to recognize these regions. According to the arguments above, the ideal value of $k$ should be as small as possible but large enough to lower the number of chaos leaves. We call this value optimization point. Obviously, different images have different optimization points. Figure \ref{fig:boundaryevlonk} shows the boundary detection Precision-Recall based on $k$ over the BSDB test set. As shown in Figure, algorithm has better Precision-Recall performances as long as $k$ increasing since larger values of $k$ give less number of chaos leaves. But after some points, there are no improvements because numbers of chaos leaves are zero. 

Based the observation of $k$, we used a program to automatically locate the optimization points for each image. The program is trivial. It pre-generates the decomposition quadtree and selects the value of $k$ which is the smallest integer in range $[3, mn]$ but holding zero number of chaos leaves. Then the rest experiments are generated based on the optimization points for each image.

\begin{figure}[!htb]
  \begin{center}
   \includegraphics[width=6in]{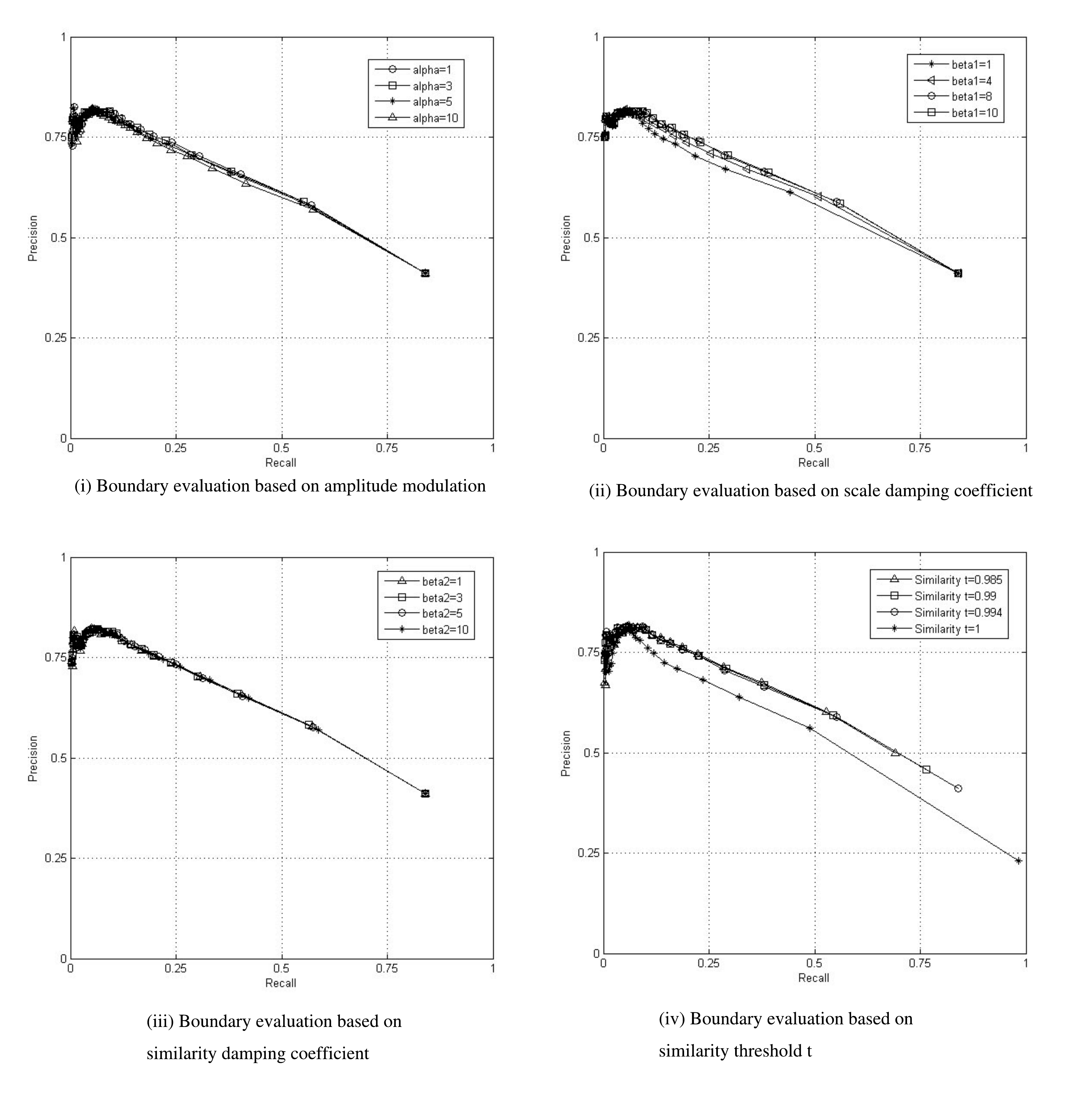}
  \end{center}
  \caption{Boundary evaluation based on $\alpha$, $\beta_1$, $\beta_2$ and $t$.} \label{fig:boundaryevlon4}
\end{figure}

\begin{figure}[!htb]
  \begin{center}
   \includegraphics[width=5in]{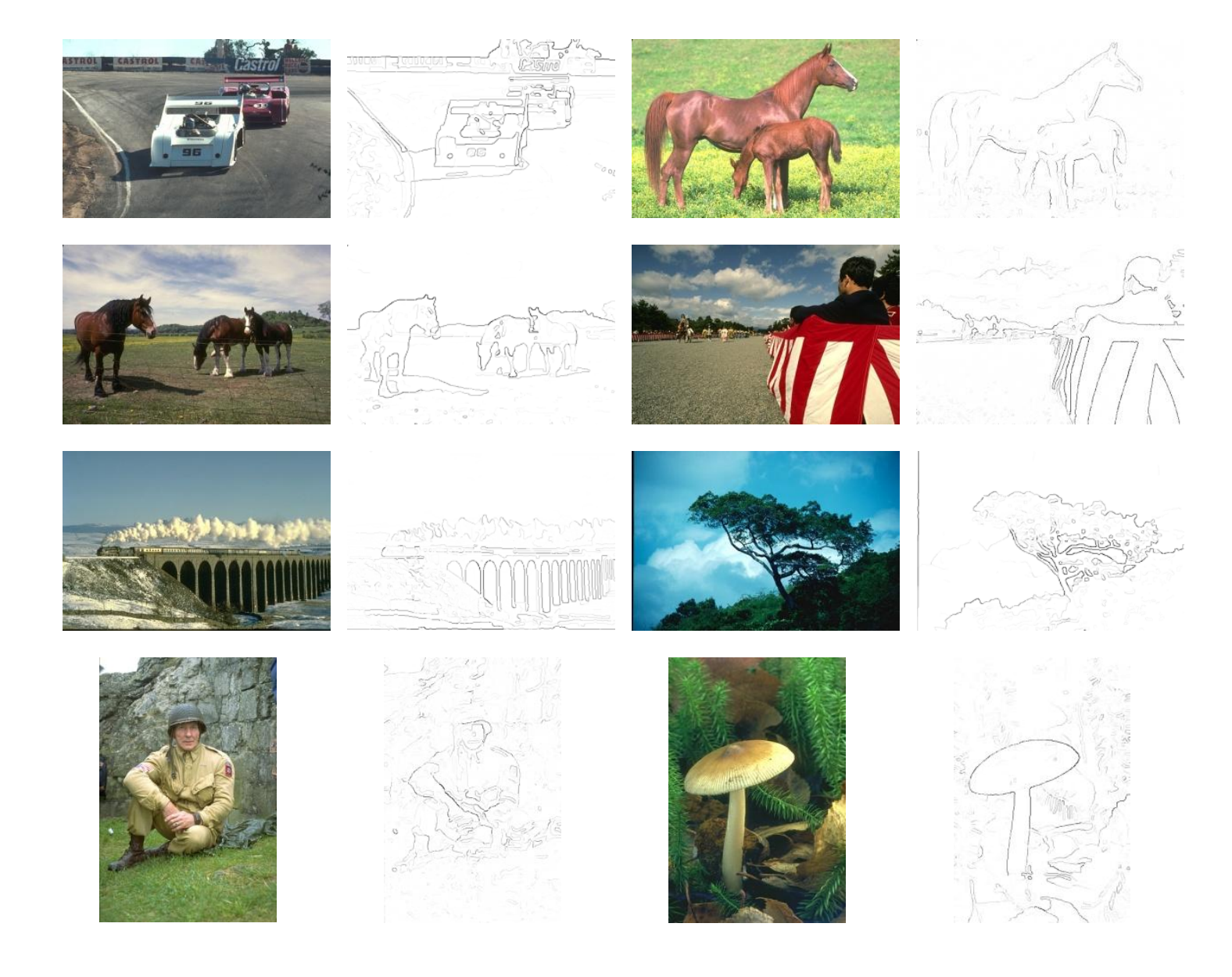}
  \end{center}
  \caption{Examples of the weighted boundaries.} \label{fig:exampleboundary}
\end{figure}

We also demonstrate the experiments based on damping coefficient $\beta_1$ of scale filter ($f_1$ \ref{fun:scalefilter}) in Figure \ref{fig:boundaryevlon4} (ii), the experiments based on damping coefficient $\beta_2$ of similarity filter ($f_2$ \ref{fun:similarityfilter}) in Figure \ref{fig:boundaryevlon4} (iii), the experiments based on amplitude modulation $\alpha$ of similarity filter ($f_2$ \ref{fun:similarityfilter}) in Figure \ref{fig:boundaryevlon4} (i) and the experiments based on similarity threshold $t$ of similarity filter ($f_2$ \ref{fun:similarityfilter}) in Figure \ref{fig:boundaryevlon4} (iv). As demonstrated, different values of these parameters slightly shift the Precision-Recall curves. Based on these experiments, we choose $\alpha=1$, $\beta_1=8$, $\beta_2=3$ and $t=0.994$ for the remaining experiments. Figure \ref{fig:exampleboundary} provides some examples of boundary detections.

\subsection{Visualization of the Segments}

\begin{figure}[!htb]
  \begin{center}
   \includegraphics[width=5in]{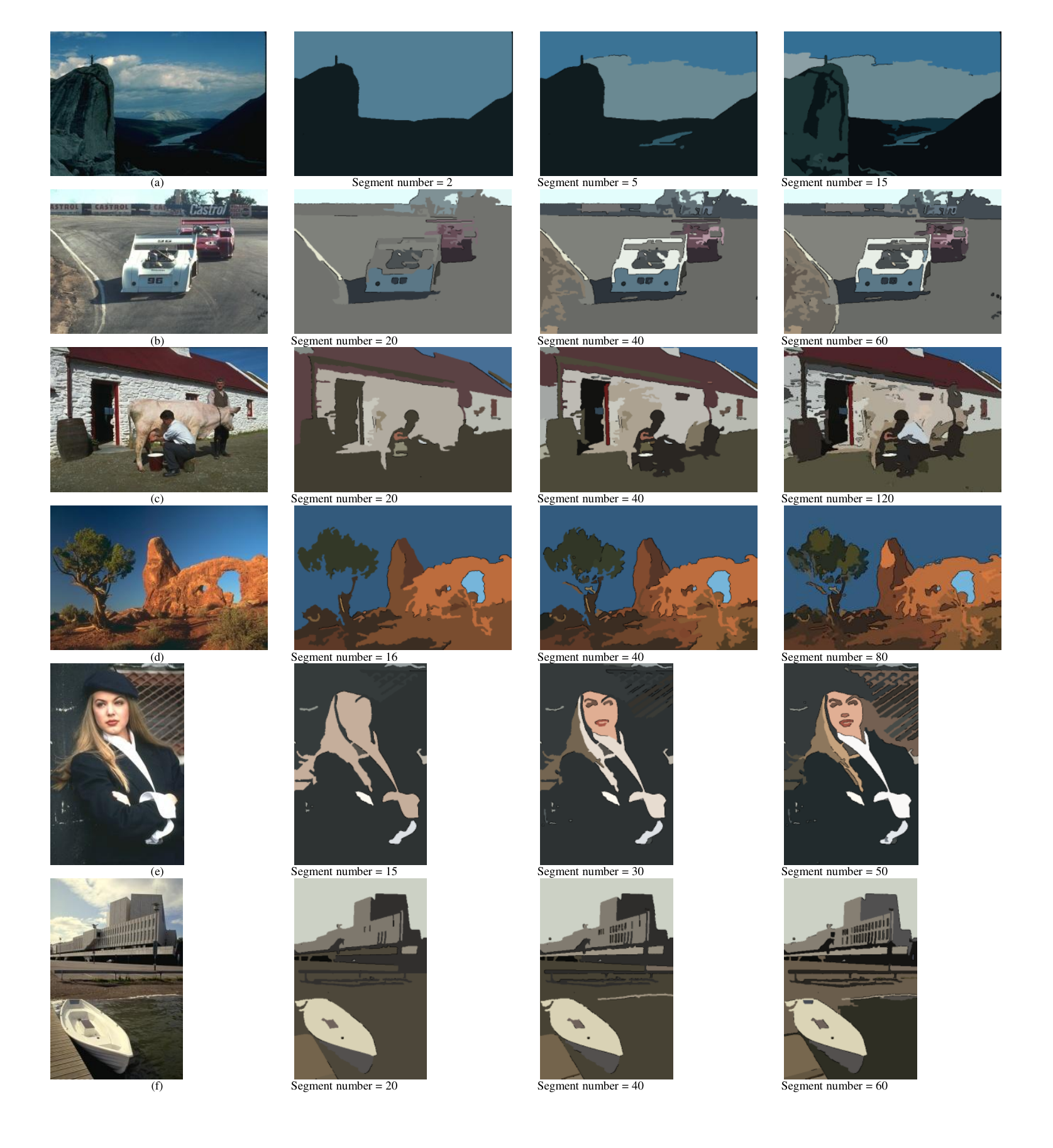}
  \end{center}
  \caption{Examples of the segmentation results by VHBS.} \label{fig:examplesegment}
\end{figure}

We demonstrate some examples of the natural images based on different number of segments in Figure \ref{fig:examplesegment}. The visualizations of the segments are generated based on top-down manner of PBT. Since PBT is constructed based on the probabilities of the segments, the segments located at the higher level of PBT suggest higher possibilities of the segments. If visualization PBT in small number of segments, it gives the general outline of the images since only the segments with high possibilities are visualized. The more details are provided when visualization goes down to the depth of PBT. 

\subsection{Evaluation Experiments by Tuning Parameters}

Rather than giving the subjective running experiments of the algorithm, the segmentation results need to be compared with other existed methods quantitatively. One usually used segmentation evaluation is supervised evaluation \cite{Chabrier2004}, which compares the results of segmentation against the manually-segmented reference images. The disadvantage of such methods is that the quality of segmentation is inherently subjective. Then, there is a requirement to evaluate the image segmentations unsubjectively. Such methods are called unsupervised evaluation \cite{Zhang2008} methods such as \cite{Zhang2004,Borsotti1998,Chen2004}. 

To evaluate the results of segmentation quantitatively and objectively, \cite{Haralick1985} proposed four criteria: (i) the characteristics should be homogeneous within the segments; (ii) the characteristics should be quite different between the adjacent segments; (iii) shape of the segments should be simple and without holes inside; (iv) boundaries between segments need to be smooth and continues, not ragged. For our experiments, we choose $Q(I)$ \cite{Borsotti1998}, $H_r(I)$, $H_l(I)$ and $E$ \cite{Zhang2004} as evaluators. $Q(I)$ and $H_r(I)$ measure the intra-region uniformity, which is described as the criteria (i). $H_l(I)$ measures the inter-region uniformity, which is the criteria (ii). $E$ is defined as $E=H_r(I)+H_l(I)$, which combine the criteria (i) and criteria (ii). Table \ref{tab:evaluators} gives the details of the evaluators used in this chapter. The disadvantage of unsupervised evaluations is that these evaluation criteria might not appropriate for the natural images since the perception of human segmentation is based on the semantic understanding. It might result different conclusions by human perception \cite{Zhang2008}.

\begin{table}
  \caption{The unsupervised evaluators used in this chapter.} 
	\label{tab:evaluators}
  \begin{tabular}{|p{4cm}|p{3cm}|l|}
    \hline
    Evaluator & Description  & Formula \\
    \hline
    $Q(I)$ \cite{Borsotti1998}    & Intra-region evaluator based on color error & $Q(I)=\frac{\sqrt{R}}{1000N\cdot M}\sum_{i=1}^{R} [\frac{e_i^2}{1+\log S_i}+(\frac{R(S_i)}{S_i})^2]$ \\ 
    \hline
    $H_r(I)$ \cite{Zhang2004}  &              & $H_r(I)=\sum_{i=1}^{R} (\frac{S_i}{S_I})H(X_i)$\\
            & Intra-region evaluator based on entropy & $H(X_i)=-\sum_{m\in V_i^\mu} \frac{L_i(m)}{S_i}\log(\frac{L_i(m)}{S_i})$\\
    \hline
    $H_l(I)$ \cite{Zhang2004}  & Inter-region evaluator based on entropy & $H_l(I)=-\sum_{i=1}^{R}  \frac{S_i}{S_I}\log(\frac{S_i}{S_I})$  \\
    \hline
    $E$  \cite{Zhang2004}      & Composite evaluator & $E=H_r(I)+H_l(I)$ \\
  	\hline
  	\multicolumn{3}{|l|}{$I$: the segmented image} \\
  	\multicolumn{3}{|l|}{$NM$: the size of the image} \\
  	\multicolumn{3}{|l|}{$R$: the number of regions in the segmented image} \\
  	\multicolumn{3}{|l|}{$S_i$: the area of pixels of the ith segment} \\
  	\multicolumn{3}{|l|}{$S_I$: the area of pixels of the image $I$} \\
  	\multicolumn{3}{|l|}{$R(S_i)$: the number of segments having the area of pixels equal to $S_i$} \\
  	\multicolumn{3}{|l|}{$e_i$: the color error in RGB space defined as } \\
  	\multicolumn{3}{|l|}{\text{    } $e_i^2=\sum_{\gamma\in\{r,g,b\}}\sum_{p\in X_i}(C_\gamma(p)-\overline{C_\gamma}(X_i))^2$} \\
  	\multicolumn{3}{|l|}{\text{    } where $C_\gamma(p)$ is the value of component $\gamma$ of pixel $p$ and $\overline{C_\gamma}(X_i)=\frac{\sum_{p\in X_i} C_\gamma (p)}{S_i}$} \\
  	\multicolumn{3}{|l|}{$V_i^\mu$: the set of all possible values associated with feature $\mu$ in segment $i$} \\
  	\multicolumn{3}{|l|}{$L_i(m)$: the number of pixels in $i$th segment that have a value of $m$ for feature $\mu$} \\
  	\hline
	\end{tabular}
\end{table}

\begin{figure}[!htb]
  \begin{center}
   \includegraphics[width=5in]{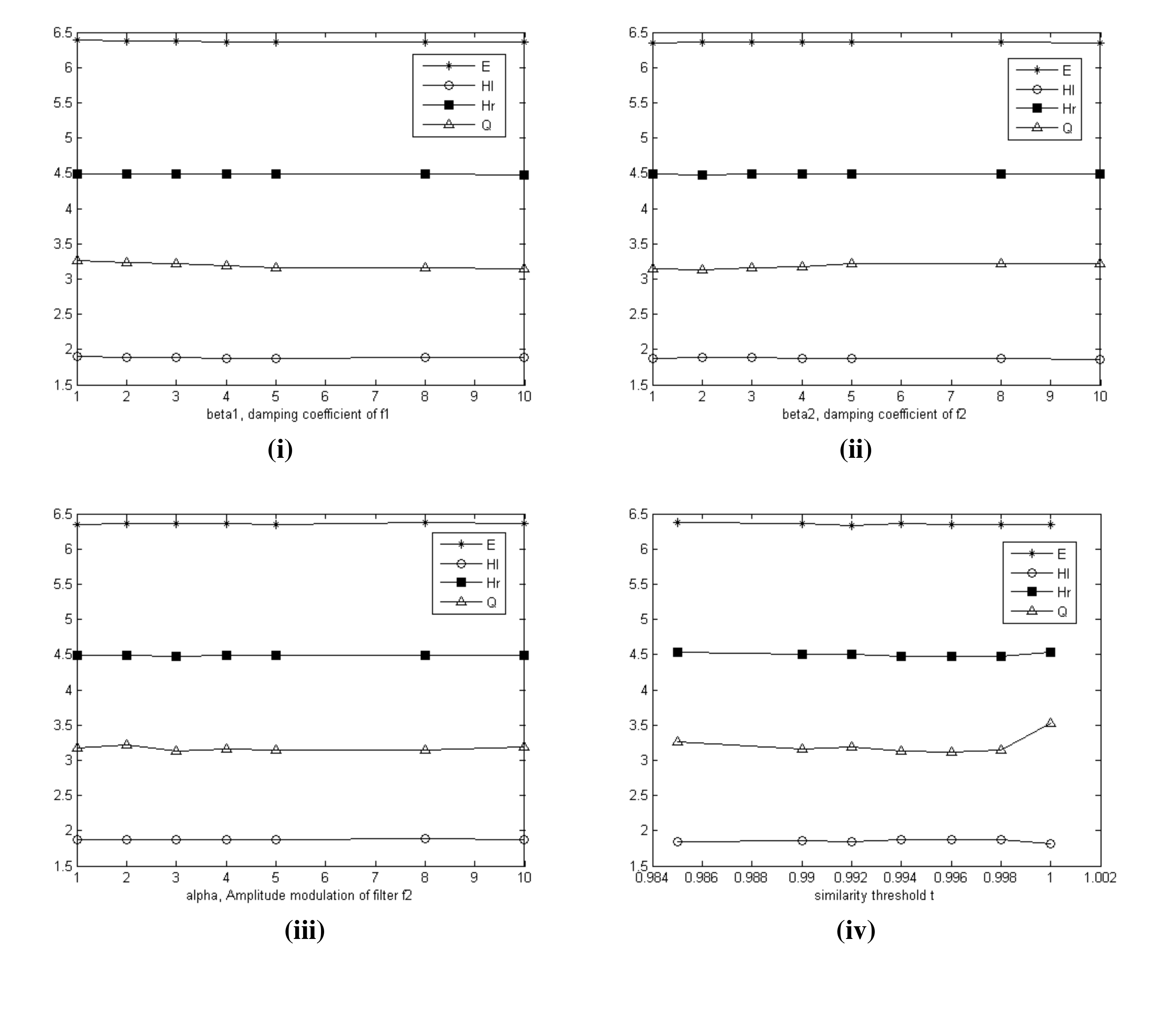}
  \end{center}
  \caption{Experimental evaluations based on $\beta_1$, $\beta_2$, $\alpha$ and $t$.} \label{fig:evaluationscoef}
\end{figure}

Figure \ref{fig:evaluationscoef} gives the experimental evaluations based on damping coefficient $\beta_1$ of scale filter $f_1$ \ref{fun:scalefilter}, damping coefficient $\beta_2$ , amplitude modulation $\alpha$ and similarity threshold $t$ of similarity filter $f_2$ \ref{fun:similarityfilter}.

Figure \ref{fig:evaluationscoef} demonstrates that the evaluations of $Q(I)$, $H_r(I)$, $H_l(I)$ and $E$ are not varied much by different damping coefficients $\beta_1$, $\beta_2$ and amplitude modulation $\alpha$. It suggests that $Q(I)$, $H_r(I)$, $H_l(I)$ and $E$ are not sensitive to parameters of $\beta_1$, $\beta_2$ and $\alpha$. Figure \ref{fig:evaluationscoef} (iv) presents the same information as Figure \ref{fig:boundaryevlon4} (iv) demonstrated that similarity threshold $t=1$ obviously has worse performance of $Q(I)$ than other similarity threshold. This is the reason we do not set $t=1$ in the experiments. 

\begin{figure}[!htb]
  \begin{center}
   \includegraphics[width=5in]{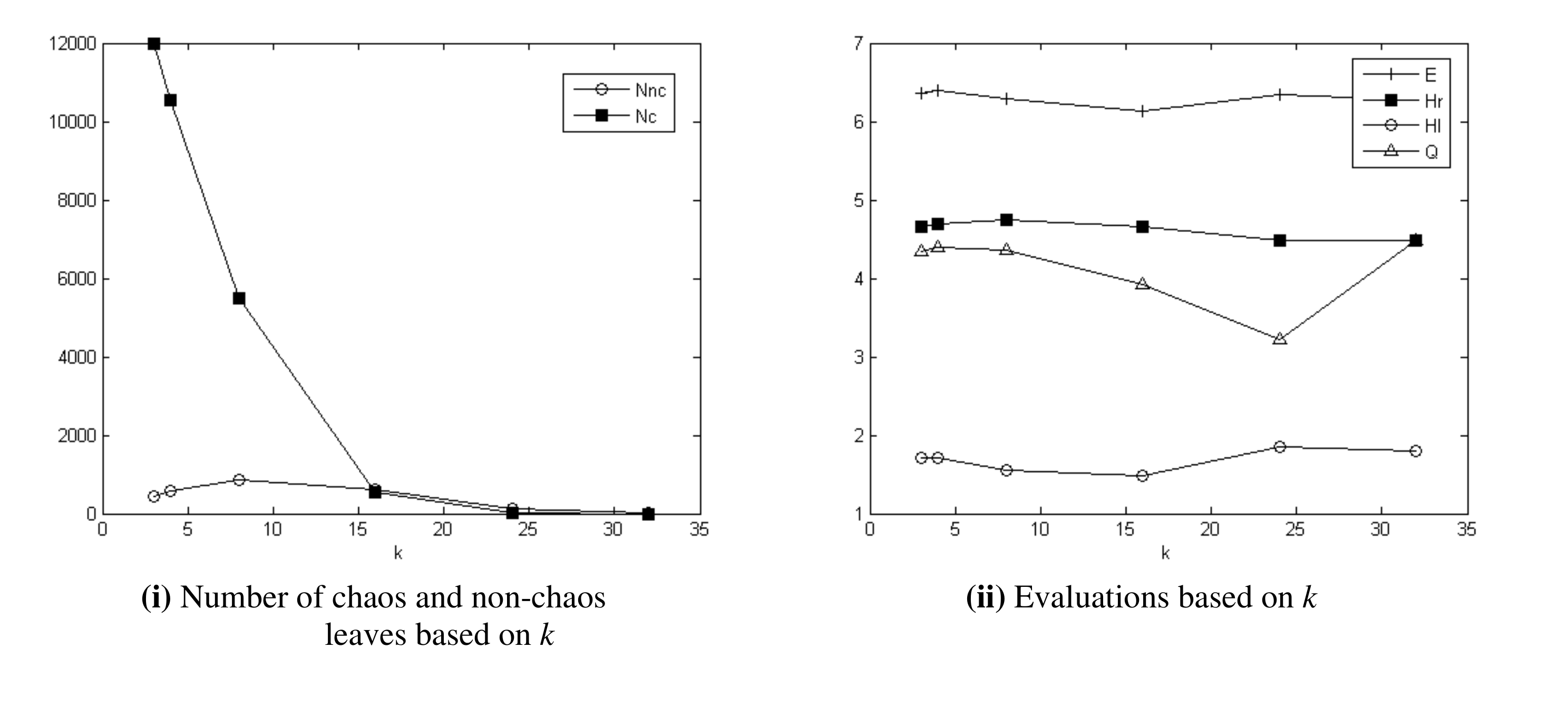}
  \end{center}
  \caption{Experimental evaluations based on $k$.} \label{fig:evaluationsk}
\end{figure}

Figure \ref{fig:evaluationsk} shows the evaluations of $Q(I)$, $H_r(I)$, $H_l(I)$ and $E$ based on parameter of $k$, where $N_{nc}$ and $N_c$ are the number of non-chaos leaves and chaos leaves. The information shown in Figure \ref{fig:evaluationsk} (i) demonstrates that both the numbers of chaos and non-chaos leaves are decreasing as long as $k$ increasing. Notice that the number of non-chaos leaves is slightly increasing at very left side of the graph is because that there are too many areas transferring from chaos to non-chaos. Based on Figure \ref{fig:evaluationsk} (i), we can make a judge that the average optimization point locates around $k=25$ for the whole test set of BSDB since $N_c$ is close to zero when $k$ is about $25$. The evaluation $Q(I)$ shown in Figure \ref{fig:evaluationsk} (ii) strongly supports this estimation. Obviously, the optimization points are different when images are different. 

\subsection{Comparison with Other Segmentation Methods}

\begin{figure}[!htb]
  \begin{center}
   \includegraphics[width=5in]{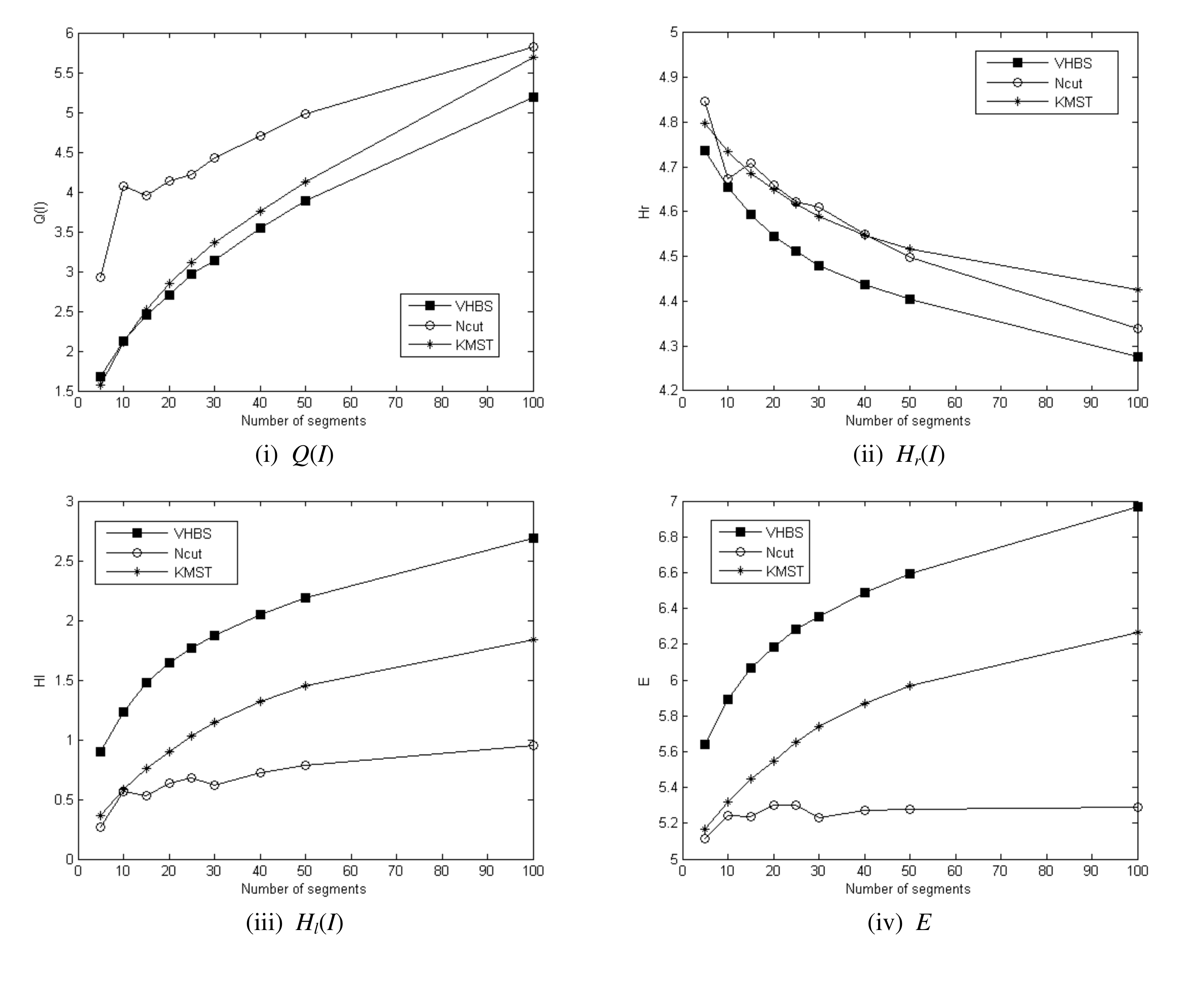}
  \end{center}
  \caption{VHBS compares with Ncut \cite{Shi1997} and KMST \cite{Felzenszwalb1998,Felzenszwalb2004}.} \label{fig:comparewithothers}
\end{figure}

In this section, we compare VHBS with Ncut \cite{Shi1997} and KMST \cite{Felzenszwalb1998,Felzenszwalb2004} over the entire test set of Berkeley Segmentation Dataset and Benchmark \cite{Martin2001}. The source codes of Ncut and KMST are got from the author's websites. To fairly compare these three algorithms, we tune the parameters to output the same number of segments of each image in the test set. Figure \ref{fig:comparewithothers} provides the evaluations of $Q(I)$, $H_r(I)$, $H_l(I)$ and $E$ based on number of segments. 

Figure \ref{fig:comparewithothers} demonstrates that our algorithm gives the best performance of $Q(I)$ and $H_r(I)$ almost over all the number of segments and Normalized Cut has the best performance of $H_l(I)$. Let us take a close look at the evaluator $H_l(I)$. As \cite{Zhang2004} points out, $H_l(I)$ favors very few large segments and many small segments. In other words, Segmentation with very few large segments and many small segments gives high evaluation $H_l(I)$.

It is expected that VHBS performs poorly over the evaluator $H_l(I)$ because $H_l(I)$ contradicts to the mechanism of scale filter $f_1$, where scale filter favors large areas. Scale filter $f_1$ \ref{fun:scalefilter} makes VHBS prefer to select large area segments, which is more consistent with human visual perception. Meanwhile, similarity filter $f_2$ \ref{fun:similarityfilter} favors the segments preserving high uniformity within the segments because $f_2$ marks the boundaries with high weights when two adjacent regions are quite different. Based on the discussion above, it is not hard to understand that VHBS performs well $Q(I)$ and $H_r(I)$, but not $H_l(I)$ and $E$ since $H_l(I)$ and $E$ is penalized by number of small area segments.

\section{Conclusion}

Our contribution lies in proposing a new low-level image segmentation algorithm, VHBS, which obeys two visual hint rules. Unlike most unsupervised segmentation methods, which are based on the clustering techniques, VHBS is based on the human perceptions since $f_1$ and $f_2$ are designed based on two visual hint rules and somehow contradict clustering ideas. The evaluations of experiments demonstrate that VHBS has high performance over the natural images. VHBS still preserves high efficiency because VHBS does not go down to the pixel level by setting the entropy and chaos thresholds as the stopping condition of the image decompositions. Rather than outputting the segments of the given images, at the same time, VHBS also provides the feature descriptors, which are statistic summarization for each segment. To improve the performance, one of our future works is to construct the algorithm in learning schema to get the optimized parameters by a learning process. 


\end{document}